%% file: FedMGP.tex
\documentclass[sigconf]{acmart}

\copyrightyear{2024}
\acmYear{2024}
\setcopyright{acmlicensed}
\acmConference[KDD '24]{Proceedings of the 30th ACM SIGKDD Conference on Knowledge Discovery and Data Mining}{August 25--29, 2024}{Barcelona, Spain}
\acmBooktitle{Proceedings of the 30th ACM SIGKDD Conference on Knowledge Discovery and Data Mining (KDD '24), August 25--29, 2024, Barcelona, Spain}
\acmDOI{10.1145/3637528.3671948}
\acmISBN{979-8-4007-0490-1/24/08}

\AtBeginDocument{%
  \providecommand\BibTeX{{%
    \normalfont B\kern-0.5em{\scshape i\kern-0.25em b}\kern-0.8em\TeX}}}

\RequirePackage{enumitem}
\setlist[itemize]{noitemsep,leftmargin=*,topsep=0em}
\setlist[enumerate]{noitemsep,leftmargin=*,topsep=0em}
\usepackage[linesnumbered,ruled,vlined,boxed,noend]{algorithm2e}
\usepackage{amsmath}  
\usepackage{multirow}
\usepackage{subfigure}
\usepackage{subcaption}
\usepackage{cleveref}
\usepackage{graphicx}
\usepackage{float}

\usepackage{balance}

\begin{document}

\title[Personalized Federated Continual Learning via Multi-granularity Prompt]{Personalized Federated Continual Learning \linebreak via Multi-granularity Prompt}

\author{Hao Yu}
\orcid{0009-0007-0705-4756}
\affiliation{%
  \institution{School of Computing and Artificial Intelligence, Southwestern University of Finance and Economics}
  \city{Chengdu}
  \country{China}
}
\email{yuhao2033@163.com}

\author{Xin Yang}
\orcid{0000-0002-0406-6774}
\authornote{Corresponding Author}
\affiliation{%
  \institution{School of Computing and Artificial Intelligence, Southwestern University of Finance and Economics}
  \city{Chengdu}
  \country{China}
}
\email{yangxin@swufe.edu.cn}

\author{Xin Gao}
\orcid{0009-0007-8265-898X}
\affiliation{%
  \institution{School of Computing and Artificial Intelligence, Southwestern University of Finance and Economics}
  \city{Chengdu}
  \country{China}
}
\email{xingaocs@hotmail.com}

\author{Yan Kang}
\orcid{0000-0002-2016-9503}
\affiliation{%
  \institution{Webank}
  \city{Shenzhen}
  \country{China}
}
\email{kangyan2003@gmail.com}

\author{Hao Wang}
\orcid{0000-0001-9492-3807}
\affiliation{%
  \institution{College of Computer Science,\\ Sichuan University}
  \city{Chengdu}
  \country{China}
}
\email{cshaowang@gmail.com}

\author{Junbo Zhang}
\orcid{0000-0001-5947-1374}
\affiliation{%
  \institution{JD Intelligent Cities Research}
  \institution{JD iCity, JD Technology}
  \city{Beijing}
  \country{China}
}
\email{msjunbozhang@outlook.com}

\author{Tianrui Li}
\orcid{0000-0003-2581-840X}
\affiliation{%
  \institution{School of Computing and Artificial Intelligence, Southwest Jiaotong University}
  \city{Chengdu}
  \country{China}
}
\email{trli@swjtu.edu.cn}

\renewcommand{\shortauthors}{Hao Yu et al.}

\begin{abstract}
  \input{text/0_abstract}
\end{abstract}



\makeatletter \gdef\@ACM@checkaffil{} \makeatother
\maketitle

\input{text/1_introduction}
\input{text/2_relatedwork}
\input{text/3_problem}
\input{text/4_method}

\input{text/5_experiments}

\input{text/6_conclusion}

\begin{acks}
This work was supported by the National Natural Science Foundation of China (Nos. 72242106, 62176221), the Natural Science Foundation of Sichuan Province (No. 2022NSFSC0528), Sichuan Science and Technology Program (No. 2024YFHZ0024), Jiaozi Institute of Fintech Innovation in Southwestern University of Finance and Economics (Nos. kjcgzh20230103, kjcgzh20230201) and the Fundamental Research Funds for the Central Universities (YJ202421).
\end{acks}

\bibliographystyle{ACM-Reference-Format}
\balance
\bibliography{ref}

\appendix
\input{text/9_Appendix}

\end{document}

%% file: text/0_abstract.tex
Personalized Federated Continual Learning (PFCL) is a new practical scenario that poses greater challenges in sharing and personalizing knowledge. PFCL not only relies on knowledge fusion for server aggregation at the global spatial-temporal perspective but also needs model improvement for each client according to the local requirements. Existing methods, whether in Personalized Federated Learning (PFL) or Federated Continual Learning (FCL), have overlooked the multi-granularity representation of knowledge, which can be utilized to overcome Spatial-Temporal Catastrophic Forgetting (STCF) and adopt generalized knowledge to itself by coarse-to-fine human cognitive mechanisms. Moreover, it allows more effectively to personalized shared knowledge, thus serving its own purpose. To this end, we propose a novel concept called multi-granularity prompt, i.e., \textit{coarse-grained global prompt} acquired through the common model learning process, and \textit{fine-grained local prompt} used to personalize the generalized representation. The former focuses on efficiently transferring shared global knowledge without spatial forgetting, and the latter emphasizes specific learning of personalized local knowledge to overcome temporal forgetting. 
In addition, we design a \textit{selective prompt fusion} mechanism for aggregating knowledge of global prompts distilled from different clients. By the exclusive fusion of coarse-grained knowledge, we achieve the transmission and refinement of common knowledge among clients, further enhancing the performance of personalization. Extensive experiments demonstrate the effectiveness of the proposed method in addressing STCF as well as improving personalized performance. Our code now is available at \href{https://github.com/SkyOfBeginning/FedMGP}{https://github.com/SkyOfBeginning/FedMGP}.

%% file: text/1_introduction.tex
\section{Introduction}
Federated Continual Learning (FCL) is a new practical paradigm aiming at fusing knowledge from different times and spaces without catastrophic forgetting in dynamic Federated Learning (FL) settings \cite{yang2023federated}. Moreover, Personalized Federated Learning (PFL) tries to fuse implicit common knowledge extracted from various clients and personalize the generalized knowledge for better performance on the client side \cite{tan2022towards}.
However, to better accommodate diverse local requirements in highly heterogeneous FCL scenarios, personalization solutions are necessary for leveraging knowledge fused from different spatial and temporal perspectives. 
Therefore, Personalized Federated Continual Learning (PFCL) is proposed as a combination of PFL and FCL with broader application scenarios. 

\begin{figure}[htbp]
    \centering
    \includegraphics[width=0.49\textwidth]{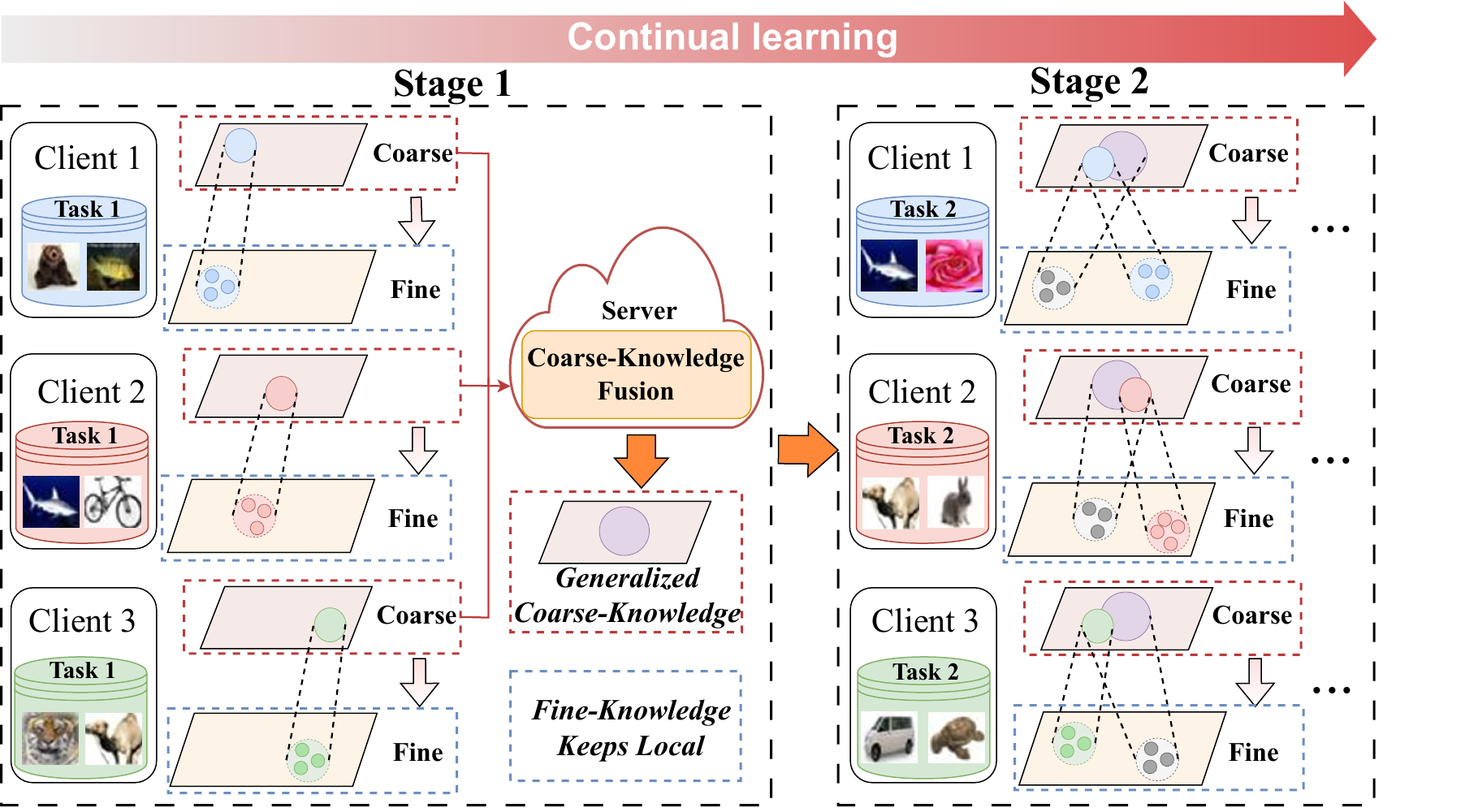}
    \caption{An illustration of constructing a multi-granularity knowledge space in PFCL. By dividing local knowledge into coarse-grained knowledge and fine-grained knowledge, better aggregation of common representations can be achieved on the server side. On the local side, fine-grained knowledge is used to personalize the generalized representation. The two different levels of knowledge can accumulate over time.}
    \label{intro}
\end{figure}

PFCL is more challenging than static PFL because of the higher requirements for handling heterogeneous knowledge. On the one hand, it implies accumulating knowledge against spatial-temporal catastrophic forgetting (STCF), which is the main issue of FCL.
On the other hand, it also needs to achieve effective extraction and fusion of client-specific and invariant knowledge, ensuring local personalization post the integration of shared knowledge. This is the primary goal of PFL.
Both issues can be solved by the multi-granularity representation of knowledge. Therefore, we can effectively address these issues by constructing a multi-granularity knowledge space, as illustrated in \autoref{intro}. Existing methods have not taken this way into account. 

The key to personalization lies in the accurate isolation of knowledge, namely, turning local knowledge into client-invariant and client-specific knowledge. By introducing a multi-granularity knowledge space, it is easy to decompose knowledge into coarse-grained, representing common aspects, and fine-grained, representing specific aspects, during the training stage to fulfill the requirements of personalization.
Moreover, on the server side, we consolidate client-invariant knowledge while maintaining the distinctiveness of client-specific knowledge with the exclusive fusion of coarse-grained representations.

For STCF, the knowledge learned by the deep network is overly fine-grained, making it highly susceptible to degradation in performance, such as the weighted average of parameters, leading to serious forgetting. In contrast, this multi-granularity knowledge representation exhibits stronger robustness against forgetting. On the one hand, using coarse-grained representation for temporal-spatial invariant knowledge makes it easier to transfer and fuse knowledge across time and space. On the other hand, utilizing fine-grained representation captures time-specific and space-specific knowledge, thereby achieving better personalization.

Inspired by the human cognitive process \cite{yang2023federated}, knowledge transfer is based on shared cognition, such as a common language. In this work, we construct a multi-granularity knowledge space by utilizing prompts of different granularity, namely \textit{coarse-grained global prompts} and \textit{fine-grained local prompts}, with a pre-trained Vision Transformer (ViT) \cite{dosovitskiy2020image}. Specifically, we employ the pre-trained ViT as the shared public cognition. We train coarse-grained prompts operating at the input without altering internal parameters to represent temporal-spatial invariant knowledge. Subsequently, leveraging the frozen coarse-grained prompts, we train class-wise fine-grained prompts that directly interact with the multi-head self-attention layer as temporal-spatial specific knowledge. This fine-tuning process enhances the model's ability to adapt to local tasks. This coarse-to-fine cognitive approach also aligns with the human cognitive process, where attention is initially directed toward outlines before focusing on details. Finally, we design a selective prompt fusion mechanism on the server side. This novel prompt fusion approach further mitigates spatial forgetting caused by aggregation. The contributions of this paper are summarized as follows:

\begin{itemize}[leftmargin=*]
  \item We formally define a new personalized federated learning scenario called PFCL, which imposes higher demands on knowledge processing while preventing spatial-temporal forgetting. For the first time, we construct a multi-granularity knowledge space in this scenario, effectively addressing these challenges.
  \item We propose a novel method called Federated Multi-Granularity Prompt (FedMGP), which introduces two distinct prompt levels to represent coarse-grained and fine-grained knowledge, respectively. It effectively overcomes STCF while meeting the requirements for personalization.
  \item Extensive experiments demonstrate that our method achieves state-of-the-art performance in two different scenarios of federated continual learning. Moreover, our approach exhibits superior performance in personalizing and retaining temporal and spatial knowledge.
\end{itemize}

%% file: text/2_relatedwork.tex
\section{Related Work}
\subsection{Multi-Granularity Computing}
Multi-granularity computing addresses the challenge of tackling the coexistence of data with different granularities \cite{yang2022three,yang2022temporal}. Extracting multi-granularity knowledge benefits our understanding of materials and their intrinsic properties. 

VL-PET \cite{hu2023vl} designs a multi-granularity controlled mechanism to impose control on modular modifications of the pre-trained language model at coarse and fine granularities. \cite{chen2023multi} constructs a question-answering dataset with yearly, monthly, and daily-grained data and proposes MultiQA to address temporally multi-granularity question-answering. \cite{yang2022three} adopts the sequential three-way decision method to extract knowledge of different granularities in open-topic classification tasks. 
\cite{xiao2018group} decouples the objects of group re-identification tasks into individual, subgroup, and entire group granularities to handle the dynamic changes in group layout and member variations. \cite{pan2021privacy} introduces granular computing in FL and achieves automatic neural architecture search to adapt the different information granularity across clients. \cite{ma2022layer} achieves a fine-grained knowledge fusion with layer-wised aggregation. PartialFed \cite{sun2021partialfed} transfers cross-domain knowledge of adaptive granularity among clients by automatically switching the learning strategy. \cite{cai2022multi} utilizes bi-directional guidance with a prior attention mechanism to transfer coarse-grained and fine-grained knowledge among multi-scale local models in an extremely heterogeneous federated system. \cite{liu2020client} proposes a hierarchical FL framework to reduce the communication overhead, conducting model aggregation at two granularities.

Currently, the core concept of multi-granularity cognition has been gradually embraced by the general public and is progressively being applied to scenarios involving spatial-temporal changes. For the traffic accident predictions, given the dynamic nature of road networks and expanding urban areas, it is challenging when the spatial-temporal granularity of forecasting improves due to the rarity of accident records and the complexity of long-term future dependencies. To address these challenges, \cite{zhou2020foresee} propose a unified framework named RiskSeq, which is designed to foresee sparse urban accidents with finer granularities and multiple steps from a spatial-temporal perspective. This approach aims to enhance the accuracy and detail of accident predictions, thereby improving the efficiency of police force allocation and traffic management strategies. For the traffic flow predictions, not only should it consider the temporal dependencies that exist between different nodes in the network, but also the spatial correlations between nodes. \cite{fang2019gstnet} propose a Global Spatial-Temporal Network (GSTNet), which is composed of multiple spatial-temporal blocks, in order to capture the global dynamic spatial-temporal correlations.

However, there is currently very limited research involving multi-granularity knowledge transfer in federated learning, and there is almost no research on using multi-granularity knowledge to address the spatial-temporal catastrophic forgetting in FCL. In this paper, we retain fine-grained knowledge in the local prompts and coarse-grained knowledge in the global prompts to achieve spatial-temporal knowledge fusion across tasks and clients.

\subsection{Prompt-Based Continual Learning}
Continual Learning (CL) aims to overcome catastrophic forgetting of the previous knowledge after training on new data in non-stationary task streams \cite{de2021continual}. Various CL techniques \cite{masana2022class,li2023cl,mai2022online} have been proposed to alleviate catastrophic forgetting and achieve knowledge transfer across tasks, including regularization, rehearsal, parameter isolation, and knowledge distillation. 

Recent works introduce prompt learning to CL to achieve more efficient exemplar-free CL \cite{wang2022learning,wang2022dualprompt,smith2023coda}. Prompt learning is a novel transfer learning technique applied to adapt general knowledge of pre-trained large language or vision models to downstream tasks by optimizing prompts \cite{lester2021power,jia2022visual,zhou2022learning,kang2023grounding}. CoOp \cite{zhou2022learning} integrates learnable prompts in the vision-language model to facilitate end-to-end learning where the design of task-specific prompts is fully automated. 
L2P \cite{wang2022learning} applies learnable task-specific prompts to mitigate forgetting and even outperforms exemplar-based methods in accuracy and efficiency. DualPrompt \cite{wang2022dualprompt} decouples the learnable prompts into general and expert prompts, encoding task-invariant and task-specific knowledge, respectively. CODA \cite{smith2023coda} replaces key-value pairs in the prompt selection strategy with an attention-based end-to-end scheme. Pro-KT \cite{li2023learning} attaches complementary prompts to a pre-trained large model to efficiently transfer task-aware and task-specific knowledge. LGCL \cite{khan2023introducing} mitigates forgetting in extremely heterogeneous task streams, where the class set of each task is disjoint, by improving the key lookup of the prompt pool and mapping the output feature to class-level language representation. 

In this paper, we design a local prompt and a global prompt mechanism to extract and encode coarse-grained and fine-grained knowledge, achieving spatial-temporal knowledge transfer.

\subsection{Personalized Federated Learning}
Personalized Federated Learning (PFL) focuses on training customized models to accommodate various preferences and requirements of clients in heterogeneous FL. Existing works on PFL can be categorized into data-based and model-based approaches \cite{tan2022towards}.

Per-FedAvg \cite{fallah2020personalized} designs a Model-Agnostic Meta-Learning (MAML) framework to find a generalized global model. It trains personalized local models derived from the shared global model. pFedMe \cite{t2020personalized} integrates L2-norm regularization in the loss function to adaptively control the balance between personalization and generalization in federated MAML. Ditto \cite{li2021ditto} adds a regularization term in the local objectives as the loss function of the local adaptation process but aggregates the models before the local adaptation to strike a balance of personalization and generalization. FedSteg \cite{yang2020fedsteg} enables domain adaptation from the shared global model to personalized local models by adding a correlation alignment layer before the softmax layer. FedPer \cite{arivazhagan2019federated,pillutla2022federated} decouples the model into base layers and personalized layers and aggregates the shallow base layers to capture generic knowledge while retaining the deep personalized layers locally to maintain personalized knowledge. FedMSplit \cite{chen2022fedmsplit} adopts multi-task learning to fit related but personalized models for clients. FedCE \cite{cai2023fedce} clusters the clients into several groups based on the similarity of local data distributions and trains multiple global models for each group. FedCP\cite{zhang2023fedcp} proposes an auxiliary Conditional Policy Network to achieve more fine-grained personalization with sample-wise feature separation.
\cite{vahidian2023efficient} conducts clustering by analyzing the principal angles of local data in the subspaces and delays the training stage until the clustering is accomplished. These works do not explicitly explore the multi-granular knowledge in the processes of generalization and personalization.

Some recent works incorporated prompting learning methods into PFL. pFedPG \cite{yang2023efficient} utilizes personalized prompt generation globally and personalized prompt adaptation locally to achieve PFL under heterogeneous data. pFedPrompt \cite{guo2023pfedprompt} extracts user consensus from linguistic space and adapts to local characteristics in visual space in a non-parametric manner.

However, extracting and fusing spatial-temporal multi-granular knowledge via prompting to overcome catastrophic forgetting and data heterogeneity has not yet been implemented in PFCL.

%% file: text/3_problem.tex
\section{Problem Definition}
\subsection{Personalized Federated Continual Learning}
\label{sec:3.1}
The primary goal of PFCL is to accumulate and fuse knowledge from different times and spaces. Clients employ suitable personalized strategies to make the received generalized knowledge better adapted to the characteristics of local data and effectively meet the requirements of local tasks. However, due to FCL itself, PFCL is also susceptible to severe spatial-temporal catastrophic forgetting. 

Therefore, PFCL has \textit{three main objectives}. The first is to form more generalized knowledge during server knowledge fusion, avoiding spatial catastrophic forgetting caused by heterogeneous data. The second is for clients to adopt appropriate strategies to overcome temporal forgetting resulting from continual learning. The third is for clients to employ suitable personalization strategies, ensuring that the received generalized global model better adapts to the local task requirements and characteristics of local data.


Now, we extend the traditional FL to PFCL. 
\begin{itemize}[leftmargin=*]
    \item Given $a$ clients (denoted as $\mathcal{A}=\{A_1, A_2, \ldots, A_a\}$), and a central server (denoted as $S$), each client $\{A_i, 1 \leq i \leq a\}$ has its unique task sequence $\mathcal{T}_i$, where each task encompasses different classes. The task sequence of client $A_i$ is denoted as $\mathcal{T}_i=\{T_i^1, T_i^2, \ldots, T_i^{n_i}\}$, where $n_i$ represents the total number of tasks on client $A_i$. The $k$-th task of $\mathcal{T}_i$ contains $\left| \mathcal{C}_i^k \right|$ classes, and $\mathcal{C}_i = \{\mathcal{C}_i^1\cup \mathcal{C}_i^2\cup \ldots, \cup \mathcal{C}_i^{n_i}\}$. 
    
    \item During the training of task $r$, the global model on the server already possesses the knowledge of $T_i^1$ to $T_i^{r-1}$ from client $\{A_i, 1 \leq i \leq a\}$. The server $S$ then distributes it back to clients. 
    After personalizing the received global model $\theta_g^{r-1}$, the client $A_i$ continually trains it on $T_i^r$ as the initial model to get the new local model $\theta_i^r$. The local model $\theta_i^r$ should perform well in classifying classes from the set $\{\mathcal{C}_i^{1} \cup \mathcal{C}_i^{2} \cup \ldots, \cup \mathcal{C}_i^r\}$.
    
    \item Finally, the server collects the local models from clients who participate in FCL and obtains a new global model $\theta_g^r$, which has more generalized knowledge of learned tasks from all clients. Clients need to adopt appropriate strategies to personalize the global model $\theta_g^r$, enabling it to perform better locally.
\end{itemize}

According to the similarity of task sequences among clients, FCL can be initially divided into two scenarios: synchronous FCL and asynchronous FCL \cite{yang2023federated}. We will discuss it in detail in \autoref{sec:5.1}.

\subsection{Spatial-Temporal Catastrophic Forgetting}
Catastrophic Forgetting is a fundamental challenge in CL, which refers to a phenomenon that a model would forget the knowledge learned on old tasks when training on new tasks \cite{de2021continual}. The reason for catastrophic forgetting is that the well-learned network parameters on the old tasks are overwritten during training on the new tasks \cite{yang2023federated}. 

In the FCL setting, catastrophic forgetting exists as well.
In a real-world scenario, data reaches clients consecutively through task streams \cite{li2024towards}, causing temporal catastrophic forgetting. At the aggregation stage, the central server collects local models and aggregates them into one global model. Then, the server distributes the global model back to clients. 
Local models are trained with different training data. Aggregating them leads to the overwriting of certain task-specific crucial parameters, consequently causing a decline in the performance of the global model on local-specific tasks.
Adopting the global model consolidated such conflict knowledge exacerbates the temporal catastrophic forgetting of each client's previous tasks.

The fundamental reason for STCF is that the knowledge represented by the model's parameters is too fine-grained, leading to a lack of robustness against minor variations. Therefore, it is necessary to represent knowledge in a multi-granularity way. Splitting it into coarse-grained spatial-temporal-invariant knowledge and fine-grained spatial-temporal-specific knowledge and handling them separately can effectively overcome STCF.

We design \textbf{Temporal Knowledge Retention} to measure the effectiveness of temporal knowledge transfer and \textbf{Spatial Knowledge Retention} to measure the effectiveness of spatial knowledge transfer in PFCL. 

\textit{Definition 1. \textbf{(Temporal Knowledge Retention)}} Given a federated learning system with $a$ clients, the temporal knowledge retention is defined as:

\begin{equation}
\label{krt}
    KR_t=\frac{1}{a}\sum_{i=1}^a\frac{Acc(\theta^r_i;T^0_i)}{Acc(\theta^0_i;T^0_i)},
\end{equation}
where $Acc(\theta^r_i;T^0_i)$ denotes the test accuracy of client $A_i$'s local model at $r$-th round on the $0$-th task and $Acc(\theta^0_i;T^0_i)$ denotes the test accuracy of client $A_i$'s local model at the initial round on the $0$-th task.

\textit{Definition 2. \textbf{(Spatial Knowledge Retention)}} Given a federated learning system with $a$ clients, the spatial knowledge retention is defined as:
\begin{equation}
\label{krs}
    KR_s=\frac{1}{a}\sum_{i=1}^a\frac{Acc(\theta^r_g;T^r_i)}{Acc(\theta^r_i;T^r_i)},
\end{equation}
where $Acc(\theta^r_g;T^r_i)$ denotes the accuracy of the global model $\theta^r_g$ on the current local task $T^r_i$ at client $A_i$ and $Acc(\theta^r_i;T^r_i)$ denotes the accuracy of the local model $\theta^r_i$ on its current local task $T^r_i$.

%% file: text/4_method.tex
\section{Multi-granularity Prompt}
\label{sec4}
In this section, we elaborate on our proposed Federated Multi-Granularity Prompt (FedMGP), which introduces a multi-granularity knowledge space into PFCL for the first time to better address personalized requirements and spatial-temporal forgetting.

Specifically, on the client, we design prompts at two granularity levels for knowledge representation, namely \textit{Coarse-grained Global Prompt} (\textbf{see \autoref{sec4.1}}) and \textit{Fine-grained Local Prompt} (\textbf{see \autoref{sec4.2}}).
Global prompts represent coarse-grained common knowledge, while local prompts, built upon global prompts, represent class-wise fine-grained knowledge. Only fusing the coarse-grained common knowledge facilitates the formation of generalized knowledge and avoids spatial forgetting caused by aggregating fine-grained knowledge. Local prompts based on global prompts aim to personalize the generalized knowledge from the server while preventing temporal forgetting due to class increments.

On the server side, we devise a new approach for fusing global prompts called \textit{Selective Prompt Fusion} (\textbf{see \autoref{sec4.3}}) without spatial forgetting. Aggregating only coarse-grained knowledge not only enhances aggregation speed but also provides further improvements in privacy protection.

\begin{figure*}[htbp]
    \centering
    \includegraphics[width=1\textwidth]{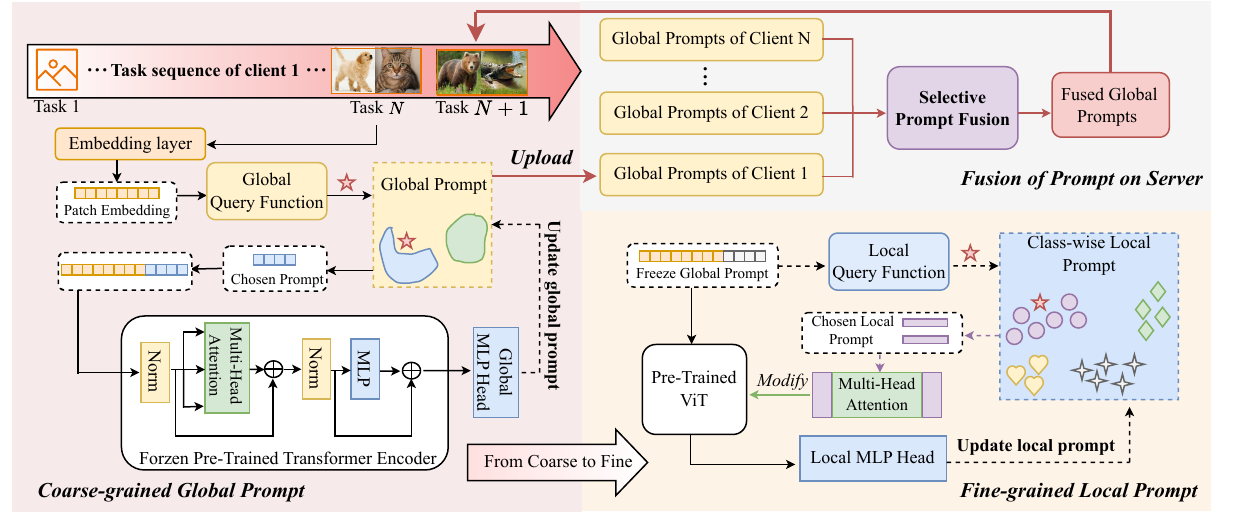}
    \caption{An overview of the proposed FedMGP. Two granularities of prompts are used to capture both temporal-spatial invariant knowledge and specific knowledge. The \textit{coarse-grained global prompt} is trained through a shared ViT model, acting on the embedding layer. The \textit{fine-grained local prompt} is built upon the coarse-grained prompt by introducing additional parameters in the MSA layer, enabling the model to better adapt to local data. Moreover, selective prompt fusion is employed to aggregate global prompts on the server side, forming generalized knowledge.}
    \label{framework}
\end{figure*}

The overall framework of the proposed method is shown in \autoref{framework}, and the algorithm is summarized in \autoref{alo1}. 

\begin{algorithm}[!htb]
\caption{FedMGP Algorithm.}
\label{alo1}
  \KwIn{$a$ clients $\mathcal{A} = \{A_i\}_{i=1}^a$ with their own task sequence $\mathcal{T}_i = \{T_i^n\}_{n=1}^N$, a pre-trained frozen ViT $\mathcal{V}$ without classification head.}
  \KwOut{Fused global prompt pool $\mathcal{P}_G$, local prompt pool ${P}_l = \{{P}_l^i\}_{i=1}^a$ of all clients and local classification head $\mathcal{}{H}_l = \{{H}_l^i\}_{i=1}^a$. }
  
  Initialization\;
  \While{task number n $\leq$ N}{
        \For{each client $A_i, 1 \leq i \leq a$}{
        $\mathcal{V}_g^i$ $\gets$ \rm{LoadGHead}(${H}_g^i,\mathcal{V}$)\;

        \rm{\textbf{Training global prompts:}}
        
        \For{each $\{x, y\} \in T_i^n$}{
        $E$ $\gets$ \rm{EmbeddingLayer}($x$)\;

        $\{{K}_g,P_g\}$ $\gets$ \rm{GlobalQueryFunction}(${P}_g^i,E,\mathcal{V}$)\; \tcp{Key-value pair.}

        $E'$ $\gets$ \rm{AppendGP}($E,P_g$)\;

        $L_g$ $\gets$ \rm{Classify}($\mathcal{V}_g^i,E',Y$) \; \tcp{Classification loss with $\mathcal{V}_g^i$.}

        \rm{Optimize}($L_g,{H}_g^i,{K}_g,P_g$)\;

        }
         \rm{\textbf{Freeze global prompts;}}
         
        \rm{\textbf{Training local prompts:}}

        \For{each $\{x, y\} \in T_i^n$}{

        $E'$ $\gets$ \rm{GetGlobalPrompt}($x,\mathcal{P}_g^i,\mathcal{V}$)\;
        
        $\{{K}_l,P_l\}$ $\gets$ \rm{LocalQueryFunction}($\mathcal{P}_l^i,E',\mathcal{V}$);

         $\mathcal{V}_l^i$ $\gets$ \rm{LoadLocalPrompt\&Head}(${H}_l^i,\mathcal{V}$,$P_l$)\;

        $L_l$ $\gets$ \rm{Classify}($\mathcal{V}_l^i,E',Y$);

        \rm{Optimize}($L_l,\mathcal{H}_l^i,{K}_l,P_l$)\;\tcp{Notice that global prompts are frozen.}
        }}  
        
        \rm{\textbf{Server aggregation:}}
        
        $\mathcal{P}_g$ = \{$\mathcal{P}_g^{1} \cup \ldots \cup \mathcal{C}_g^{T_a}$ \} \;

        $\mathcal{P}_G$ $\gets$ \rm{SelectivePromptFusion}($\mathcal{P}_g$) \;

        Distribute $\mathcal{P}_G$ to all clients for next task training.
    }
\end{algorithm}

\subsection{Coarse-grained Global Prompt}
\label{sec4.1}
Due to the heterogeneity of data, significant differences exist among local models, leading to substantial variations in extracted knowledge. This also poses significant challenges for the fusion and transfer of knowledge, as the knowledge learned by each client is overly fine-grained. Inspired by the cognitive processes of humans, knowledge transfer among humans is effective because there is a fundamental shared cognition, enabling the meaningful exchange of knowledge. Therefore, we assign each client with the same pre-trained ViT model as a foundational cognitive system. With ViT’s parameters frozen, clients learn global prompts that operate at the input level. Consequently, these global prompts represent coarse-grained knowledge acquired through the common model learning process. Furthermore, as the knowledge is extracted from the same model, it is more convenient to aggregate knowledge on the server side without spatial forgetting.

The training of coarse-grained global prompts is based on the frozen ViT model. Moreover, global prompts operate at the input level, not influencing the model's parameters. The purpose is to extract knowledge into a common space through the same model. 

\subsubsection{Global Prompt Pool}
Taking inspiration from L2P \cite{wang2022learning}, we devise a prompt pool for storing and selecting the global prompts. The prompt pool is defined as 
\begin{equation}
    \mathcal{P}_g=\{P_g^1, P_g^2, \ldots, P_g^M\},
\end{equation}
where $m$ is the pool size and $P_g^j$ is a single global prompt. Then, let $x$ and $E = f_e(x)$ be the input and its corresponding embedding feature, respectively. Denoting $\{s_i\}_1^N$ be the indices of $N$ global prompts, then we can modify the embedding feature as follows:
\begin{equation}
    E' = \left[ P_g^{s_1}, \ldots, P_g^{s_k} ; E\right], 1 \leq N \leq M,
\end{equation}
where [;] represents concatenation along the token length dimension. The next question is how to choose global prompts.

\subsubsection{Global Query Function}
Due to the use of the same model, similar inputs tend to select similar prompts and vice versa. This mitigates the challenge of aggregating heterogeneous knowledge on the server. Based on this, we have designed a key-value pair-based query strategy that dynamically selects suitable prompts by calculating the similarity between the input key and existing prompts' keys.

We associate each prompt in the pool with a learnable key, denoted as $\{(K_g^1,P_g^1),(K_g^2,P_g^2),\ldots,(K_g^m,P_g^m)\}$. To ensure that similar inputs have similar keys, we use the output features of the pre-trained ViT $\mathcal{V}$ as the key for the input, i.e., $K_g^{in} = \mathcal{V}(E)$. Then, the query process can be summarized by the following expression:
\begin{equation}
    \mathcal{K}^s_g = \underset{\mathcal{K}_g}{\operatorname{argmin}} \sum_{i=1}^{N} \text{dis}(K_g^{in}, K_g^{i}),
\end{equation}
where $\mathcal{K}_s$ denotes the subset of top-N keys selected specifically for the input, and $\mathcal{K}_g$ represents the set of keys for all global prompts. In this work, we utilize cosine similarity as the distance function to measure the similarity between keys. 

\subsubsection{Optimization for Global Prompt}
Each client has a global classification head used for training global prompts, denoted as $H^i_g$. At the beginning of training, it is necessary to load the pre-trained model with $H^i_g$ to enable it to perform the classification task, and we denote the model with $H^i_g$ as $\mathcal{V}^i_g$. Overall, the training loss function is as follows:
\begin{equation}
    \underset{H^i_g,\mathcal{P}_g,\mathcal{K}_g}{\operatorname{min}} \mathcal{L}(\mathcal{V}^i_g(E'),y)+ \lambda_1 \sum_{\mathcal{K}^{s}_g} \operatorname{dis}(K_g^{in},K_g^{s_i}),
\end{equation}
where $\lambda_1$ is a hyperparameter. The initial term comprises the softmax cross-entropy loss, while the subsequent term serves as a surrogate loss aimed at bringing selected keys closer to their corresponding query features. 


\subsection{Fine-grained Local Prompts}
\label{sec4.2}
Once the training of global prompts is completed, they will be frozen and remain unchanged, including both the prompts themselves and their corresponding keys, until the next task training. Based on the frozen global prompts, we further developed fine-grained class-wise local prompts. These prompts directly impact the model's multi-head self-attention (MSA) \cite{vaswani2017attention} layers, facilitating the extraction of local, fine-grained knowledge. Additionally, this fine-grained prompting helps overcome temporal forgetting induced by class increments. The hierarchy of prompts, from coarse to fine, simplifies generalized knowledge extraction, fusion, and personalization.

\subsubsection{From Coarse to Fine}
Similarly, a prompt pool is constructed for local prompts. However, since it represents class-specific knowledge, the size of the pool depends on the number of data classes. The local prompt pool is defined as
\begin{equation}
    \mathcal{P}_l=\{(K_l^1,P_l^1), (K_l^2,P_l^2), \ldots, (K_l^C,P_l^C)\},
\end{equation}
where $C$ represents the number of classes. It is precisely the class-wise fine-grained knowledge that imparts significant effectiveness to our approach of personalization and addressing temporal forgetting induced by class increments. It is proved in \autoref{sec5.3}.

\subsubsection{Local Query Function}
Fine-grained prompts are selected based on the global prompt, so we must first obtain frozen global prompts by allowing the original input $x$ to undergo the global query function. Subsequently, we concatenate to form an input $E'$ with selected global prompts. Then, similar to obtaining the key for global prompts, we acquire the key for local prompts $K_g^{in} = \mathcal{V}(E')$, with the only difference being that the input is now $E'$. The subsequent steps of calculating similarity and selection are analogous to the corresponding operations for global prompts.

Note that we do not employ this querying function during the \textit{training phase}. Instead, we use mask code to select the local prompt corresponding to the data class for training.

\subsubsection{Optimization for Local Prompt}
Local prompts directly operate on the model's MSA layer, where we represent the input query, key, and values as $h_q,h_k,h_v$, respectively. MSA layers can be denoted as:
\begin{equation}
    \text{MSA}(h_q,h_k,h_v) = \operatorname{Concat}\left(\mathrm{h}_{1}, \ldots, \mathrm{h}_{\mathrm{z}}\right) W^{O},
\end{equation}
where $h_i = \operatorname{Attention}({h}_{Q} W_{i}^{Q}, {h}_{K} W_{i}^{K}, {h}_{V} W_{i}^{V})$. $W$ is the project matrix and $z$ is the number of MSA layers. We use the \textbf{Prefix Tuning (Pre-T)} to tune local prompts. Pre-T splits the local prompt $P_l$ into $p_K$ and $p_V$, and adds them to $h_K$ and$h_V$:
\begin{equation}
    \operatorname{MSA}' = \operatorname{MSA}(h_Q,\left[ p_V ; h_K\right],\left[ p_V ; h_V\right]).
\end{equation}

Once global prompts have completed training, they freeze along with their corresponding keys. The input $x$ first goes through the global query function to find the corresponding global prompts. Subsequently, the embedding of $x$ is concatenated with the prompts to form $E'$. Then $E'$ is processed by the local query function to find the corresponding fine-grained prompts $\{K_l,P_l\}$. Thus, ViT modifies its MSA layer based on $P_l$ and loads the local classification head 
$H^i_l$, forming $\mathcal{V}^i_l$. The local prompt training loss function is 
\begin{equation}
    \underset{H^i_l,\mathcal{P}_l,\mathcal{K}_l}{\operatorname{min}} \mathcal{L}(\mathcal{V}^i_l(E'),y)+ \lambda_2 \sum_{\mathcal{K}^{s}_l} \operatorname{dis}(K_l^{in},K_l^{s_i}).
\end{equation}

\subsection{Selective Prompt Fusion}
\label{sec4.3}
To fuse global prompts precisely, we devise a novel \textit{selective prompt fusion} mechanism that aggregates prompts from different prompt pools through knowledge distillation, enhancing their generalization. To our knowledge, it is a novel approach to distill prompts from different clients.

We denote the small proxy dataset owned by the server as $\mathcal{D}_s$. $\{x_s,y_s\}$ are the samples and corresponding labels from $\mathcal{D}_s$ for the distillation process. For the convenience of writing and understanding, we will only consider two global prompt pools here, denoted as $\mathcal{P}_g^i$ and $\mathcal{P}_g^j$. $\mathcal{P}_g^i$ is chosen as the student pool. Initially, the input $x_p$ searches for the corresponding global prompt within $\mathcal{P}_g^i$, and then concatenates to form an embedding $E'_i$ with the prompt. Similarly, $E'_j$ represents the embedding of the same input but concatenated with the prompts from $\mathcal{P}_g^j$. Therefore, the distillation loss can be summarized as:
\begin{equation}
    \mathcal{L}_{CE} = \underset{x_s \in \mathcal{D_s}}{\operatorname{MSE}}(\mathcal{V}(E'_i),\mathcal{V}(E'_j)).
\end{equation}

%% file: text/5_experiments.tex
\section{Experiments}
\label{sec:5}
\subsection{Experimental Setup}
\subsubsection{Datasets}

\begin{table*}[htbp]
\centering
\caption{Accuracy of the aggregated global model on local test sets CIFAR-100 with 5 class-incremental tasks each client. }
\begin{tabular}{cccccccccccccc}
\toprule
\multirow{3}{*}{Algorithm} & \multirow{3}{*}{Backbone} &\multicolumn{6}{c}{Asynchronous} & \multicolumn{6}{c}{Synchronous} \\
 &  &  \multicolumn{6}{c}{Task ID} & \multicolumn{6}{c}{Task ID} \\
 &  \, \, & 1& 2 & 3 & 4 & 5 & Average & 1& 2 & 3 & 4 & 5 & Average\\

\midrule
FedAvg\cite{mcmahan2017communication} &\multirow{4}{*}{ResNet-18}   & 47.39 & 62.63 &	67.25 &	62.69 &	68.72 &	61.74  & 67.77 & 77.72 &	76.65 &	74.59 &	82.00 &	75.74 \\

FedProx\cite{li2020federated} & & 68.26 & 56.94 &	65.20 &	63.82 &		67.48 &	64.34 & 43.07 &	23.09 &	51.41 &	45.01 &	50.62 &	42.64 
 \\

FedEWC & & 27.77 &	 20.66 & 25.70 & 24.11 & 26.41 & 24.93 & 49.94 & 71.00 & 70.22 & 70.09 &	 77.89  & 67.83

 \\
GLFC\cite{dong2022federated}  &    & 14.22 &	18.84 &	23.93 &	26.70 &	22.52 &	21.24  & 5.22 & 8.93  & 24.61 &	 35.43 & 42.33 & 23.30

 \\

\midrule
FedViT & \multirow{3}{*}{ViT}& 83.02 & 82.39 & 83.24 &	80.34&	83.32 &82.46 &  70.30 &	71.05 &	69.10 &	64.36 &	71.01 &	69.16 
\\
FedL2P & &  89.63  &  89.68 &	90.45	& 90.02 & 90.75 & 90.11 
   &   80.22   &	82.81 &	 81.61 & 80.68 & 84.14 & 81.89 \\
FedDualP &  & 82.09 & 81.17 & 80.05	& 80.52 & 81.48 & 81.06  & 63.84 & 65.62 &	63.16 &	61.10 &	63.28 & 63.40  \\
\midrule
Ours(FedMGP) & \multirow{3}{*}{ViT}& \textbf{90.26} & \textbf{90.14}	& \textbf{91.29} & \textbf{90.30} & \textbf{90.83} & \textbf{90.56} & \textbf{82.23} & \textbf{84.14} & \textbf{82.01} & \textbf{82.47} & \textbf{86.44} &	\textbf{83.46}\\
Ours-w/oLP & & 88.35 &	89.19 &	 89.91 & 89.16 & 90.20 & 89.36 & 79.80 & 
 82.04 &	 79.71 & 80.20 & 83.56 & 81.06\\
Ours-w/oGP& & 86.93 & 88.52 & 82.85  & 84.11  & 87.29 & 85.94 & 78.73 & 78.92 & 77.74 &	75.16 & 79.10 & 77.93 \\

\bottomrule
\end{tabular}
\label{tab:cifar100}
\end{table*}
We conduct extensive experiments on CIFAR-100 \cite{krizhevsky2009learning} with 5 incremental tasks to evaluate the effectiveness of our FedMGP in addressing the challenges of PFCL. CIFAR-100 is a widely used benchmark dataset and consists of 60,000 RGB color images, each of size 32x32 pixels, classified into 100 different classes. We consider two practical scenarios of FCL, namely synchronous FCL and asynchronous FCL.

In the \textbf{synchronous FCL} settings \cite{yang2023federated}, clients have the same task sequences but a varied proportion of samples from each class. It is a common setting employed in existing FCL works \cite{dong2022federated}. The degree of data heterogeneity in this scenario is controlled with the Dirichlet parameter, which is set to be $1$ in our experiments. Specifically, we first partition the dataset into 5 tasks, each containing 20 classes, with no overlapping class between tasks. Then, within each task, the samples of each class are randomly divided into the same number of subsets as the total number of clients, ensuring that the data among clients is also non-overlapping.

In the \textbf{asynchronous FCL} settings \cite{yang2023federated}, some of the classes are accessible to all clients while others are private to certain clients, which is derived from pathological Non-IID in static FL \cite{mcmahan2017communication}. In this setting, we consider that there are $15$ private classes for each client. Each task contains $8$ classes. To be specific, each client first selects 15 classes unique to itself, and only that client has access to the full data of these classes. Therefore, there are 25 classes lefted as public classes shared by all clients. As a result, each client has data for 40 classes. The client then randomly divides these 40 classes into 5 tasks, each containing 8 classes.

\subsubsection{Baselines and Backbones}
\label{sec:5.1}
We compare FedMGP with FedAvg \cite{mcmahan2017communication}, FedEWC \cite{kirkpatrick2017overcoming}, FedProx \cite{li2020federated} and GLFC \cite{dong2022federated} on ResNet-18 \cite{he2016deep}. Since our method is based on ViT-B/16, we also conduct experiments on ViT-B/16 to compare FedMGP with FedViT. FedViT is a naive combination of FedAvg and ViT \cite{dosovitskiy2020image}, which performs federated training by locally updating and globally aggregating the parameters of the classifier heads iteratively. FedL2P and FedDualP are the adapted versions of two effective prompt-based methods in traditional CL, L2P \cite{wang2022learning} and DualPrompt \cite{wang2022dualprompt}, making them more suitable for use in a federated environment. More detailed descriptions are in \autoref{appendix:baseline}.

\subsubsection{Implementation Details}
In our setup, the federated system consists of five clients and one central server, and each client possesses a sequence of five tasks. We repeat experiments with three random seeds (42,1999,2024) and report the averaged outcomes. Across all methods, we fix the number of clients at five and the interval rounds for increments at five. We employ Adam as the optimizer with a learning rate of $0.001$. The whole training process is performed sequentially on an NVIDIA GPU RTX-3090.

\subsection{Expermental Results}
We use the accuracy of the aggregated global model on local test sets as the metric in \autoref{tab:cifar100}. To examine the impact of different backbone networks on the experimental results, we employed baseline methods based on two backbones, namely ResNet-18 and the pre-trained ViT.

Surprisingly, all methods generally perform better in the asynchronous setting than in the synchronous setting. This is attributed to the fact that in the synchronous setting, each task involves 20 classes. GLFC, FedAvg, and FedProx failed in both asynchronous and synchronous FCL. As expected, methods based on ViT outperformed those based on ResNet-18 in both scenarios. But FedAvg performs even better than FedViT and FedDual in synchronous FCL. This indicates that in scenarios with similar data distributions, FedAvg has the ability to challenge large pre-trained models.

In all methods using ViT as the backbone, FedL2P with prompts performed better than using ViT alone. Unfortunately, FedDualP performed even worse than the simple FedViT. We believe this is due to the heterogeneity in the learned parameters across clients. Moreover, the performance of these methods did not show significant improvement after aggregation. In fact, FedViT experienced a decrease of 3.9\% in average accuracy after aggregation in synchronous FCL and a decrease of 7.27\% in asynchronous FCL. 

Our method achieved the best performance in both synchronous and asynchronous settings, with accuracies of 90.56\% and 83.46\%, showing the state-of-the-art performance of fusing heterogeneous knowledge. Although our method performs well on this metric even without some components, such as Ours-w/oGP achieving 89.36\% and 81.06\%, and Ours-w/oLp achieving 87.29\% and 77.93\%, the ability to retain spatial-temporal knowledge is significantly affected. In the following section (\autoref{sec5.3}), we will evaluate each method using new metrics, i.e., temporal knowledge retention and spatial knowledge retention, to evaluate the resistance of spatial-temporal catastrophic forgetting.


\subsection{Ablation Studies}
\label{sec5.3}
\begin{figure*}[htbp]
    \centering   
    \subfigure[Asynchronous $KR_s$] 
        {
		\begin{minipage}[b]{.23\linewidth} 
				\centering
				\includegraphics[scale=0.36]{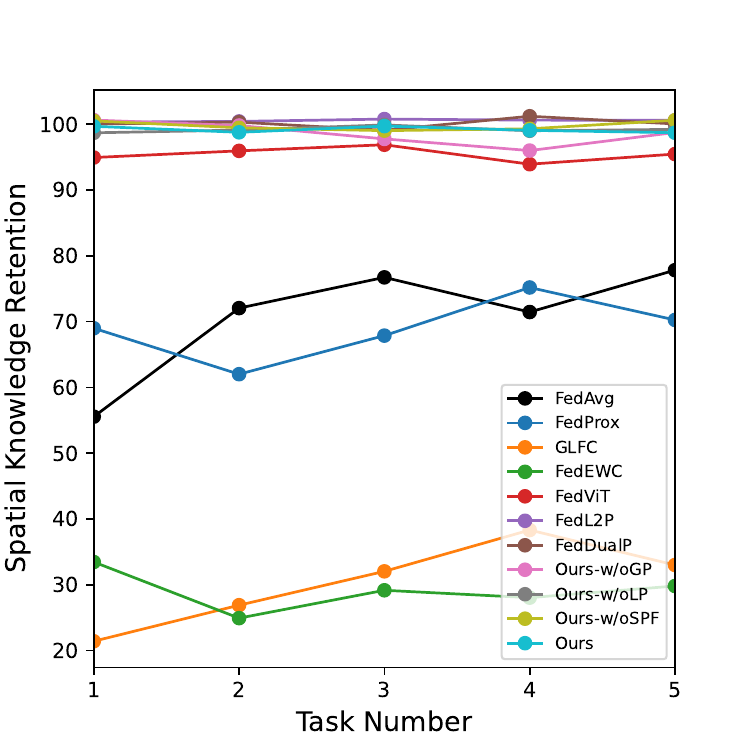}
				\end{minipage}
                \label{fig_akrs}
			}
   \hfill
		\subfigure[Asynchronous $KR_t$]
		{
			\begin{minipage}[b]{.23\linewidth}
				\centering
				\includegraphics[scale=0.36]{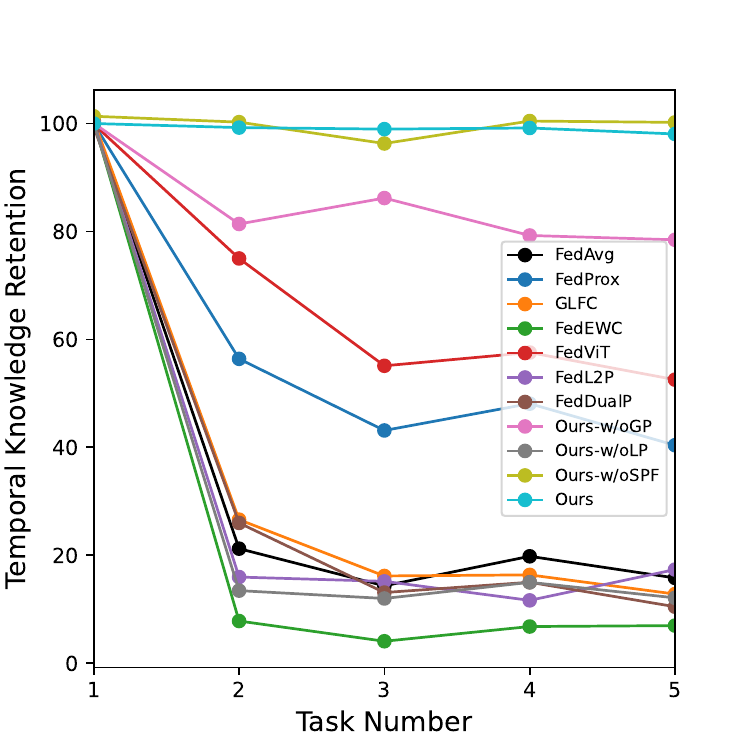}
				\end{minipage}
                \label{fig_akrt}
		}
  \subfigure[Synchronous $KR_s$]
		{
			\begin{minipage}[b]{.23\linewidth}
				\centering
				\includegraphics[scale=0.36]{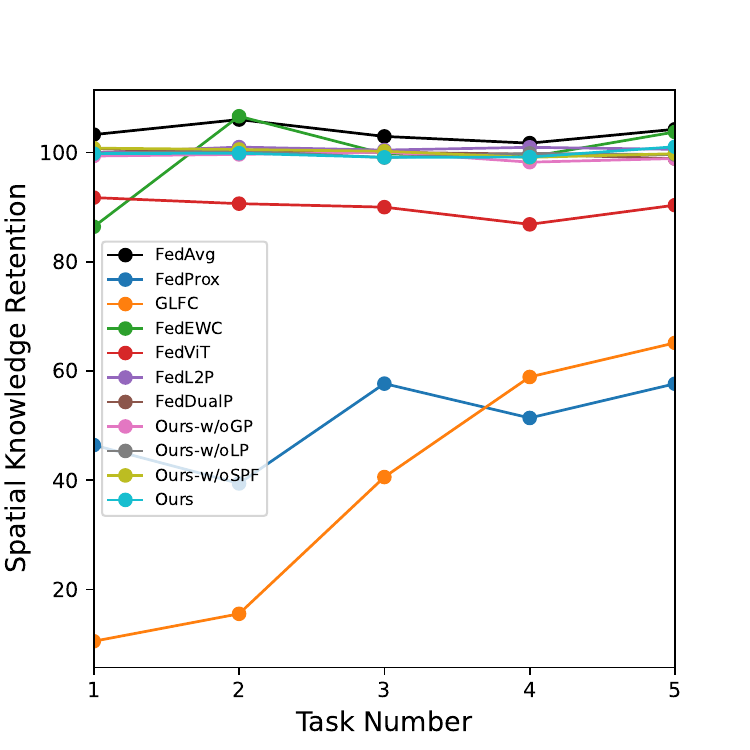}
				\end{minipage}
                \label{fig_skrs}
		}
    \subfigure[Synchronous $KR_t$]
		{
			\begin{minipage}[b]{.23\linewidth}
				\centering
				\includegraphics[scale=0.36]{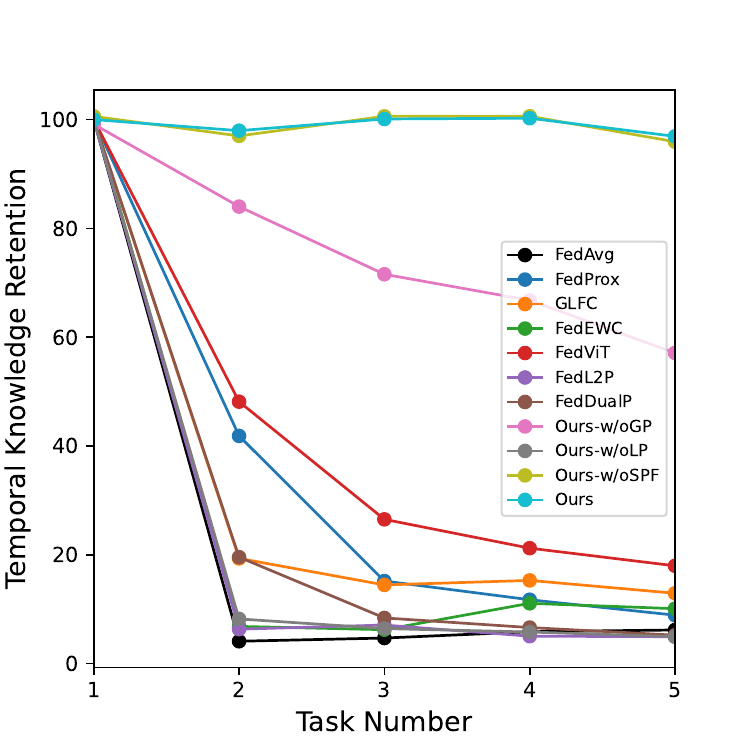}
				\end{minipage}
                \label{fig_skrt}
		}
   \caption{Ablation Studies using temporal knowledge retention (\autoref{krt}) spatial knowledge retention (\autoref{krs}) in two FCL setting. Note that, "Ours-w/oGP" refers to our method without global prompts representing coarse-grained knowledge. "Ours-w/oLP" refers to our method without local prompt to capture client/time relevant knowledge. And "Ours-w/oSPF" is to use FedAvg to aggregate global prompts instead of Selective Prompt Fusion.}
   \label{ablation}
\end{figure*}

To further validate the effectiveness of the multi-granularity knowledge space, we conducted three different ablation experiments under the same experimental setup. These experiments respectively removed the global prompts, local prompts, and the selective prompt fusion mechanism on the server. Results are shown in \autoref{ablation}.

In both asynchronous and synchronous settings, ViT-based methods have demonstrated exceptional performance in retaining spatial knowledge. This result also confirms our hypothesis: having similar cognition is the foundation for knowledge sharing. Based on that, the increment of spatial knowledge retention of FedAvg in the synchronous setting is not difficult to understand, as similar data contributes to the similarity of convolutional layers. While these methods have effectively preserved spatial knowledge, none of them demonstrates resistance to temporal catastrophic forgetting. In \autoref{fig_akrt} and \autoref{fig_skrt}, it is challenging to distinguish the difference between FedL2P, FedDualP and other methods with ResNet18 as the backbone network, as their temporal knowledge retention rates are all around 20\%. 

Our approach not only competes with other ViT methods in terms of spatial knowledge retention but also achieves almost no forgetting in temporal knowledge retention, thanks to the construction of the multi-granularity knowledge space. To evaluate the contribution of the coarse-grained global prompt and the fine-grained local prompt, three different ablation experiments are conducted, which respectively removed global prompts (Ours-w/oGP), local prompts (Ours-w/oLP), and selective prompt fusion (Ours-w/oSPF). In \autoref{fig_akrs}, there is a slight decrease in spatial knowledge retention when we remove the global prompt. The other two components have little impact on spatial forgetting. However, things become more complex when it comes to temporal knowledge retention. Without local prompts, it drops significantly to around 15\%. And when we remove global prompts, although the retention also decreases, it is not as drastic. 

It concludes that fine-grained local prompts play a crucial role in preventing temporal catastrophic forgetting, and they still need to be combined with coarse-grained knowledge to better prevent spatial-temporal catastrophic forgetting and achieve personalization. Hence, multi-granularity knowledge representation is a promising direction in PFCL.

\subsection{Sensitivity Analysis}
\begin{figure}[htbp]
    \centering   
    \subfigure[Global $KR_s$] 
        {
		\begin{minipage}[c]{.40\linewidth} 
				\centering
				\includegraphics[scale=0.33]{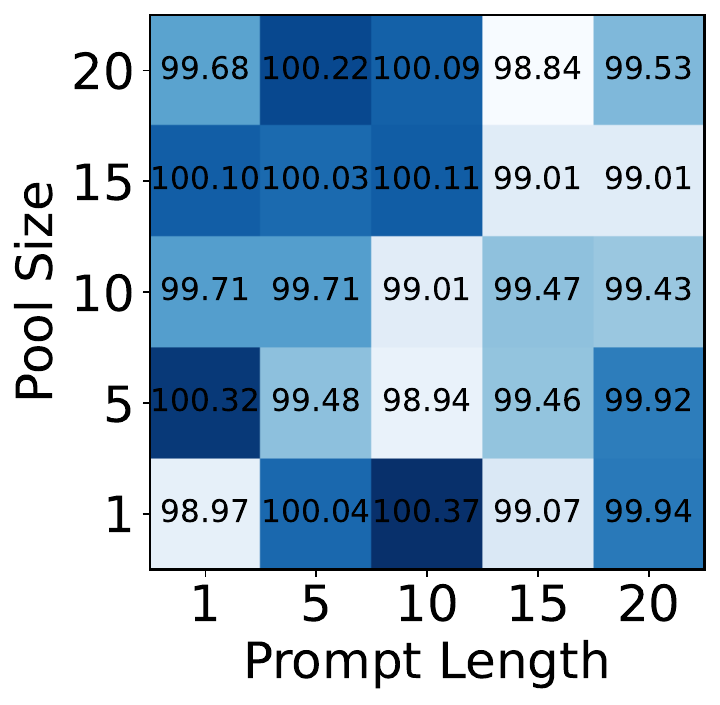}
				\end{minipage}
                \label{fig_krsg}
			}
   \hfill
		\subfigure[Local $KR_s$]
		{
			\begin{minipage}[c]{.45\linewidth}
				\centering
				\includegraphics[scale=0.33]{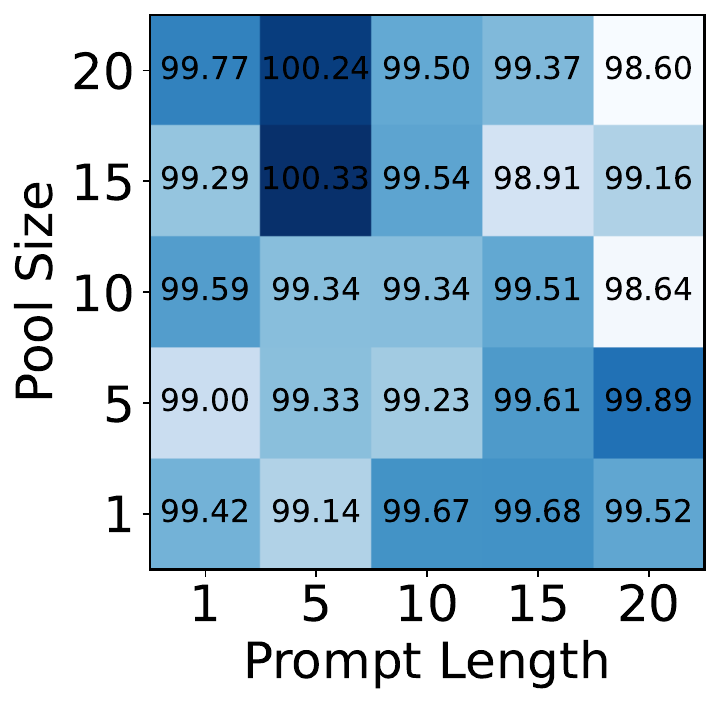}
				\end{minipage}
                \label{fig_krsl}
			}
			\caption{Sensitivity analyses of prompt length and prompt pool size. Left: Global Spatial Knowledge Retention Ratio (\%) w.r.t. prompt length $L$ and prompt pool size $M$. Right: Local Spatial Knowledge Retention Ratio (\%) w.r.t. prompt length $L$ and prompt pool size $M$.}
   \label{fig_sensitivity}
\end{figure}

FedMGP involves several hyperparameters, including prompt length, prompt pool size and so on. To further investigate the robustness of FedMGP, we conduct sensitivity analyses of prompt length and pool size on CIFAR-100 with 5 incremental tasks and present the results in \autoref{fig_sensitivity}. 

From \autoref{fig_krsg}, it can be observed that, regardless of the values of prompt length and pool size, it is beneficial for spatial knowledge of the global prompts. Additionally, under the condition of Pool Size=1 and Prompt Length=10, spatial knowledge retention is the highest, reaching 100.37\%. 

From \autoref{fig_krsl}, it can be seen that different values of prompt pool size and prompt length have little effect on spatial knowledge retention of the local prompts. It implies utilizing multi-granularity prompts is capable of training a generalized global model as well as personalized local models. More sensitivity analyses are shown in \autoref{appendix:sensitivity}.

%% file: text/6_conclusion.tex
\section{Discussion}
This section will provide a preliminary analysis and discussion of the computational cost, communication overhead, and privacy protection in federated learning for FedMGP.

\noindent\textbf{Computational cost.} The clients have only two parts to train: coarse-grained global prompts and fine-grained local prompts. The size of the global prompt pool of one client is determined by the number of prompts, prompt length, and embedding dimension, which are set to 10, 10, and 768 in the experiments. And the size of prompt keys is determined by the pool size and embedding dimension. In our experimental setup, the total size of local prompts is 4,608,000, and the size of their corresponding keys is also the same as the global prompts' keys, which is 7,680. In summary, \textit{each client has a total of 4,700,160 parameters to train}.

Moreover, the server only needs to aggregate the global prompts. This means that the training process of local prompts can proceed in parallel with the server's aggregation process.

\noindent\textbf{Communication overhead.} Our method transmits only coarse-grained global prompts and keys, keeping communication overhead low. The size of the global prompt pool per client is determined by the number of prompts, prompt length, and embedding dimension (set to 10, 10, and 768 in experiments). Prompt keys size depends on pool size and embedding dimension. Thus, the total transmitted size is 76,800 + 7,680 parameters. Although there are fine-grained local parts that also need to be trained, they remain local, which significantly reduces the communication overhead compared to traditional methods, indirectly enhancing privacy.

\noindent\textbf{Privacy protection.} Since FedMGP only transmits the coarse-grained global prompts obtained from the ViT and their keys, without uploading the original embeddings of the images and the fine-grained local prompts, FedMGP has strong privacy protection, especially against gradient leakage attacks. Moreover, in our experimental setup, the size of global prompts is only 76,800 parameters, containing much less information, which also ensures privacy protection.

\section{Conclusion}
\label{sec:discusion}
Personalized Federated Continual Learning is a novel and practical scenario. It not only requires the accumulation of knowledge that evolves over time and space but also needs consideration of personalized strategies to make generalized knowledge better adapted to local requirements. Moreover, spatial-temporal catastrophic forgetting is also a key issue that needs to be addressed.

In this paper, we first formulated a formal problem definition for PFCL and shaped the objectives of PFCL as three folds: (1) Alleviating spatial knowledge catastrophic forgetting caused by data heterogeneity; (2) Mitigating temporal knowledge catastrophic forgetting caused by dynamic task streams; (3) Training customized local models to achieve personalization. 

To address these issues, we proposed a multi-granularity knowledge space for federated continuous learning (termed as FedMGP), which has efficient fusion and personalization by representing knowledge at different granularities. Specifically, the FedMGP utilizes a shared ViT to construct coarse-grained global prompts and modifies the ViT with local prompts based on these global prompts. In addition, we designed 1) global prompts on the embedding layer to learn coarse-grained knowledge continually and 2) local prompts on the multi-head self-attention layer to learn fine-grained knowledge as a complementary to achieve personalization. Extensive experiments under synchronous and asynchronous FCL settings are conducted to demonstrate the effectiveness of our method. 

The effectiveness of multi-granularity knowledge representation has been experimentally proven in this work, and their complementary nature significantly enhances the model's resistance to spatial-temporal catastrophic forgetting. Our future research will investigate the multi-granularity representation of knowledge in various federated learning scenarios such as vertical federated learning~\cite{liu2024vfl} and multi-objective federated learning~\cite{kang2023optimizing}. We will explore its implications for privacy preservation, model performance, algorithm efficiency, and so on, aiming at achieving trustworthy PFCL.

%% file: text/9_Appendix.tex
\newpage
\appendix
\renewcommand\thesection{\arabic{section}}
\section{Notation}
\label{appendix:notation}
In \autoref{tab:notation}, we introduce the notations in our paper.
\begin{table}[h]
    \centering
      \caption{Mathematical notations and descriptions.}
    \begin{tabular}{c|c}
    \hline
        Notation & Description \\
        \hline
         $\mathcal{A}_i$& Client $i$\\
         $\theta^r_g$ & The global model at round $r$\\
         $\theta^r_i$& The local model of client $i$ at round $r$\\
         $\mathcal{P}_G$& Global prompt pool\\
         $P^i_l$& Local prompt pool of client $i$ \\
         $\mathcal{T}_i$& Task sequence of client $i$ \\
         $T^n_i$&  The task of client $i$ at incremental state $n$\\
         $\mathcal{V}$& The pre-trained ViT\\
         $E$& Embedding layer\\
         $\mathcal{H}$& The classification head \\
         \hline
    \end{tabular}
    \label{tab:notation}
\end{table}

\section{Sensitivity Analysis}
\label{appendix:sensitivity}

\begin{figure}[h]
    \subfigure[Gain Af G] 
        {
		\begin{minipage}[b]{.45\linewidth} 
				\centering
				\includegraphics[scale=0.33]{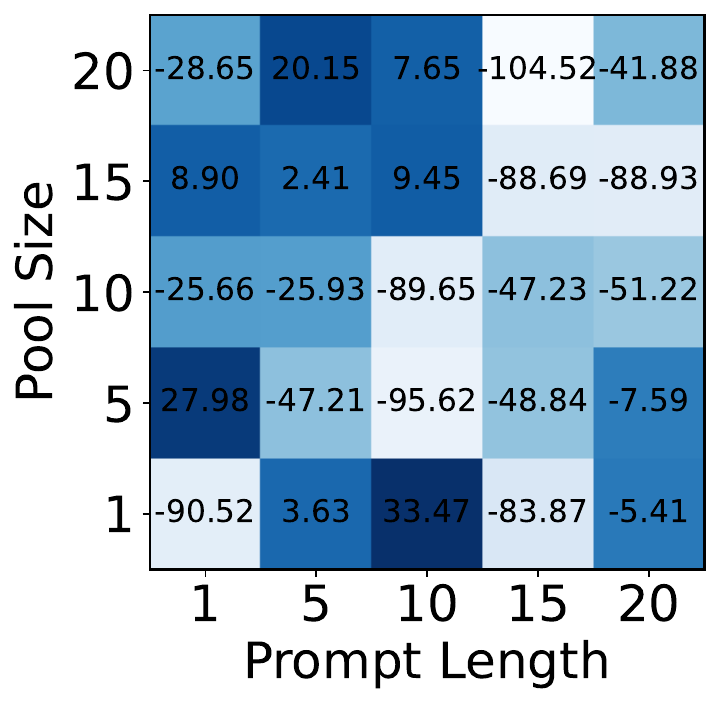}
				\end{minipage}
                \label{fig_afg}
			}
   \hfill
		\subfigure[Gain Af L]
		{
			\begin{minipage}[b]{.45\linewidth}
				\centering
				\includegraphics[scale=0.33]{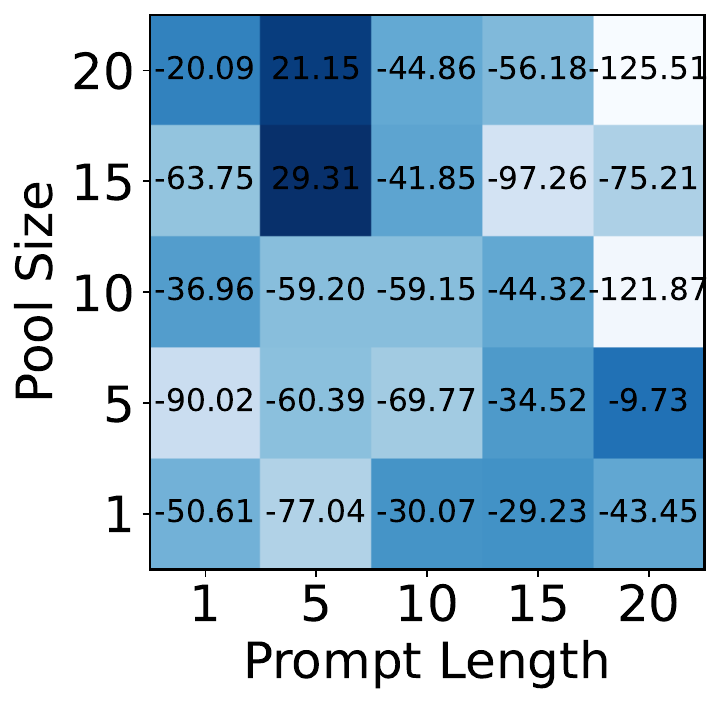}
				\end{minipage}
                \label{fig_afl}
			}
			\caption{Sensitivity analyses of prompt length and prompt pool size. Left: The performance improvement (\%) of coarse-grained global prompts after the aggregation of global prompts w.r.t. prompt length $L$ and prompt pool size $M$. Right: The performance improvement (\%) of fine-grained local prompts after the aggregation of global prompts w.r.t. prompt length $L$ and prompt pool size $M$.}
   \label{fig_sensitivity_1}
		\end{figure}

As illustrated in \autoref{fig_sensitivity_1}, the aggregation of global prompts has improved the performance of both global prompts and local prompts. \autoref{fig_afg} shows the performance improvement of coarse-grained global prompts evaluated with test accuracy (\%) after the aggregation of global prompts. \autoref{fig_afl} illustrates the performance improvement of fine-grained local prompts after the aggregation of global prompts. We can find that both improvements are robust to different values of prompt pool size and prompt length.

\begin{figure}[h]
    \subfigure[Global $KR_t$] 
        {
		\begin{minipage}[b]{.45\linewidth} 
				\centering
				\includegraphics[scale=0.33]{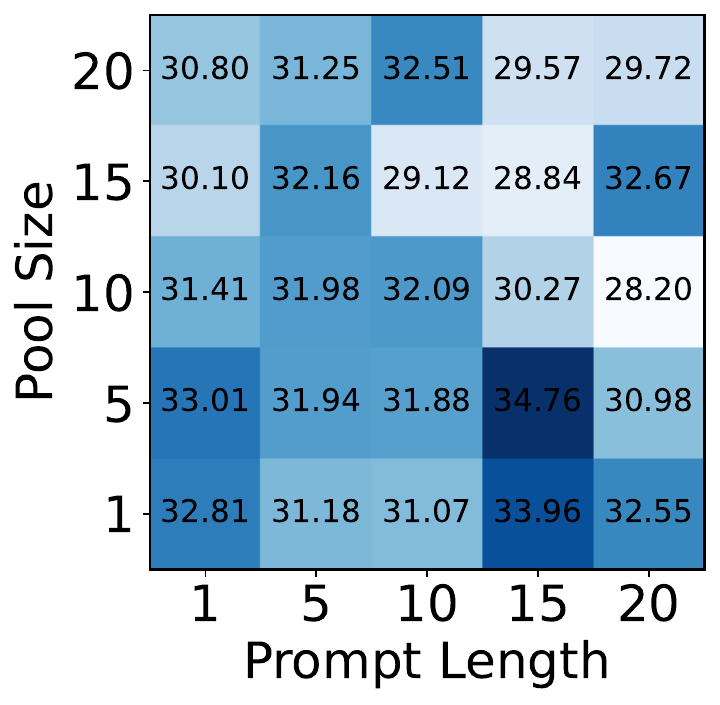}
				\end{minipage}
                \label{fig_krtg}
			}
   \hfill
		\subfigure[Local $KR_t$]
		{
			\begin{minipage}[b]{.45\linewidth}
				\centering
				\includegraphics[scale=0.33]{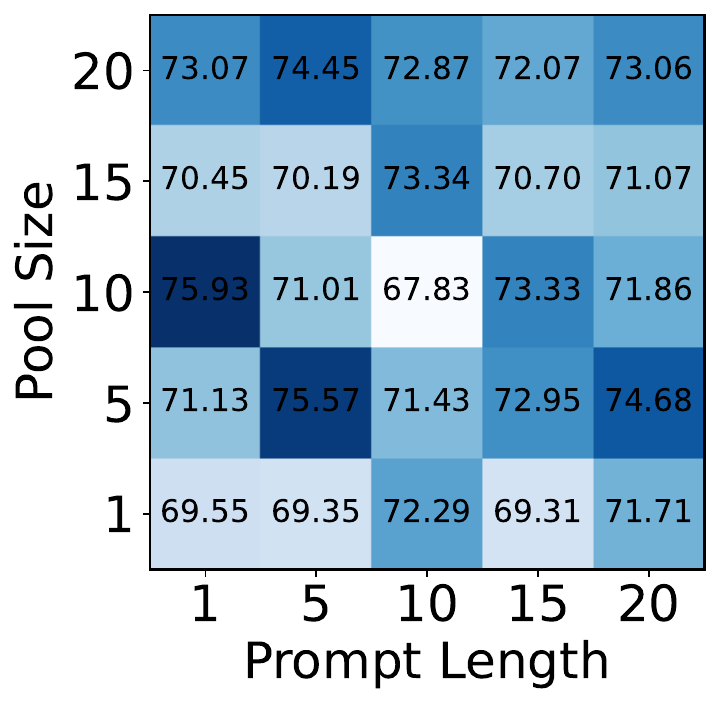}
				\end{minipage}
                \label{fig_krtl}
			}
			\caption{Sensitivity analyses of prompt length and prompt pool size. Left: Global Temporal Knowledge Retention Ratio (\%) w.r.t. prompt length $L$ and prompt pool size $M$. Right: Local Temporal Knowledge Retention Ratio (\%) w.r.t. prompt length $L$ and prompt pool size $M$.}
   \label{fig_sensitivity_2}
		\end{figure}

\autoref{fig_sensitivity_2} illustrates the temporal knowledge retention of the global model and local models. As illustrated in the left sub-figure \autoref{fig_krtg}, the performance of global spatial knowledge retention exhibits robustness to different values of prompt length and prompt pool size, which means FedMGP achieves spatial-temporal transfer effectively. From \autoref{fig_krtl}, we can conclude that different values of prompt pool size and prompt length have little effect on local spatial knowledge retention, indicating that FedMGP mitigates temporal catastrophic forgetting. It implies utilizing multi-granularity prompts is capable of training a generalized global model as well as personalized local models.

\section{Baselines}
\label{appendix:baseline}
\textbf{FedAvg} \cite{mcmahan2017communication}: FedAvg is a fundamental algorithm in federated learning. It works by first distributing a global model to multiple clients. Each client trains the model locally using its own data for a few epochs. Then, the clients send their locally updated models back to a central server. The server aggregates these local models by computing their weighted average to update the global model. This process is repeated for several rounds until the global model converges.\\
\textbf{FedEWC} \cite{kirkpatrick2017overcoming}: a combination of FedAvg and EWC, which is a commonly used baseline in PFL and FCL. EWC is a regularization-based CL method, mitigating forgetting by penalizing the changes of important parameters of the previous tasks.\\
\textbf{FedProx} \cite{li2020federated}: a heterogeneous and static FL method. It smooths data heterogeneity by adding a proximal term in the local objective.\\
\textbf{GLFC} (Glocal Local Forgetting Compensation) \cite{dong2022federated}: a synchronous FCIL method. GLFC designs a class-aware gradient compensation loss and a class-semantic relation distillation loss to mitigate forgetting and distill consistent inter-class relations across tasks. A proxy server is implemented to select the optimal previous global model to assist the class-semantic relation distillation and a prototype gradient-based communication mechanism is developed to protect data privacy.\\
\textbf{FedViT} \cite{dosovitskiy2020image}: a hybrid method of ViT and FedAvg. The global aggregation is performed by computing the average weights of the classification heads.\\
\textbf{FedL2P} \cite{wang2022learning}: a hybrid method of L2P and FedAvg. L2P is a prompt-based CL method, which applies learnable task-specific prompts to mitigate forgetting.\\
\textbf{FedDualP} \cite{wang2022dualprompt}: a hybrid method of DualPrompt and FedAvg. DualPrompt, a prompt-based CL method derived from L2P, decouples the learnable prompts into general and expert prompts, encoding task-invariant and task-specific knowledge, respectively.

%% file: FedMGP.bbl

\begin{thebibliography}{50}


\ifx \showCODEN    \undefined \def \showCODEN     #1{\unskip}     \fi
\ifx \showDOI      \undefined \def \showDOI       #1{#1}\fi
\ifx \showISBNx    \undefined \def \showISBNx     #1{\unskip}     \fi
\ifx \showISBNxiii \undefined \def \showISBNxiii  #1{\unskip}     \fi
\ifx \showISSN     \undefined \def \showISSN      #1{\unskip}     \fi
\ifx \showLCCN     \undefined \def \showLCCN      #1{\unskip}     \fi
\ifx \shownote     \undefined \def \shownote      #1{#1}          \fi
\ifx \showarticletitle \undefined \def \showarticletitle #1{#1}   \fi
\ifx \showURL      \undefined \def \showURL       {\relax}        \fi
\providecommand\bibfield[2]{#2}
\providecommand\bibinfo[2]{#2}
\providecommand\natexlab[1]{#1}
\providecommand\showeprint[2][]{arXiv:#2}

\bibitem[Arivazhagan et~al\mbox{.}(2019)]%
        {arivazhagan2019federated}
\bibfield{author}{\bibinfo{person}{Manoj~Ghuhan Arivazhagan}, \bibinfo{person}{Vinay Aggarwal}, \bibinfo{person}{Aaditya~Kumar Singh}, {and} \bibinfo{person}{Sunav Choudhary}.} \bibinfo{year}{2019}\natexlab{}.
\newblock \showarticletitle{Federated learning with personalization layers}.
\newblock \bibinfo{journal}{\emph{arXiv preprint arXiv:1912.00818}} (\bibinfo{year}{2019}).
\newblock


\bibitem[Cai et~al\mbox{.}(2023)]%
        {cai2023fedce}
\bibfield{author}{\bibinfo{person}{Luxin Cai}, \bibinfo{person}{Naiyue Chen}, \bibinfo{person}{Yuanzhouhan Cao}, \bibinfo{person}{Jiahuan He}, {and} \bibinfo{person}{Yidong Li}.} \bibinfo{year}{2023}\natexlab{}.
\newblock \showarticletitle{FedCE: Personalized Federated Learning Method based on Clustering Ensembles}. In \bibinfo{booktitle}{\emph{Proceedings of the 31st ACM International Conference on Multimedia}}. \bibinfo{pages}{1625--1633}.
\newblock


\bibitem[Cai et~al\mbox{.}(2022)]%
        {cai2022multi}
\bibfield{author}{\bibinfo{person}{Shangxuan Cai}, \bibinfo{person}{Yunfeng Zhao}, \bibinfo{person}{Zhicheng Liu}, \bibinfo{person}{Chao Qiu}, \bibinfo{person}{Xiaofei Wang}, {and} \bibinfo{person}{Qinghua Hu}.} \bibinfo{year}{2022}\natexlab{}.
\newblock \showarticletitle{Multi-granularity Weighted Federated Learning in Heterogeneous Mobile Edge Computing Systems}. In \bibinfo{booktitle}{\emph{2022 IEEE 42nd International Conference on Distributed Computing Systems}}. IEEE, \bibinfo{pages}{436--446}.
\newblock


\bibitem[Chen and Zhang(2022)]%
        {chen2022fedmsplit}
\bibfield{author}{\bibinfo{person}{Jiayi Chen} {and} \bibinfo{person}{Aidong Zhang}.} \bibinfo{year}{2022}\natexlab{}.
\newblock \showarticletitle{FedMSplit: Correlation-adaptive federated multi-task learning across multimodal split networks}. In \bibinfo{booktitle}{\emph{Proceedings of the 28th ACM SIGKDD Conference on Knowledge Discovery and Data Mining}}. \bibinfo{pages}{87--96}.
\newblock


\bibitem[Chen et~al\mbox{.}(2023)]%
        {chen2023multi}
\bibfield{author}{\bibinfo{person}{Ziyang Chen}, \bibinfo{person}{Jinzhi Liao}, {and} \bibinfo{person}{Xiang Zhao}.} \bibinfo{year}{2023}\natexlab{}.
\newblock \showarticletitle{Multi-granularity Temporal Question Answering over Knowledge Graphs}. In \bibinfo{booktitle}{\emph{Proceedings of the 61st Annual Meeting of the Association for Computational Linguistics}}. \bibinfo{pages}{11378--11392}.
\newblock


\bibitem[De~Lange et~al\mbox{.}(2021)]%
        {de2021continual}
\bibfield{author}{\bibinfo{person}{Matthias De~Lange}, \bibinfo{person}{Rahaf Aljundi}, \bibinfo{person}{Marc Masana}, \bibinfo{person}{Sarah Parisot}, \bibinfo{person}{Xu Jia}, \bibinfo{person}{Ale{\v{s}} Leonardis}, \bibinfo{person}{Gregory Slabaugh}, {and} \bibinfo{person}{Tinne Tuytelaars}.} \bibinfo{year}{2021}\natexlab{}.
\newblock \showarticletitle{A continual learning survey: Defying forgetting in classification tasks}.
\newblock \bibinfo{journal}{\emph{IEEE Transactions on Pattern Analysis and Machine Intelligence}} \bibinfo{volume}{44}, \bibinfo{number}{7} (\bibinfo{year}{2021}), \bibinfo{pages}{3366--3385}.
\newblock


\bibitem[Dong et~al\mbox{.}(2022)]%
        {dong2022federated}
\bibfield{author}{\bibinfo{person}{Jiahua Dong}, \bibinfo{person}{Lixu Wang}, \bibinfo{person}{Zhen Fang}, \bibinfo{person}{Gan Sun}, \bibinfo{person}{Shichao Xu}, \bibinfo{person}{Xiao Wang}, {and} \bibinfo{person}{Qi Zhu}.} \bibinfo{year}{2022}\natexlab{}.
\newblock \showarticletitle{Federated class-incremental learning}. In \bibinfo{booktitle}{\emph{Proceedings of the IEEE/CVF Conference on Computer Vision and Pattern Recognition}}. \bibinfo{pages}{10164--10173}.
\newblock


\bibitem[Dosovitskiy et~al\mbox{.}(2020)]%
        {dosovitskiy2020image}
\bibfield{author}{\bibinfo{person}{Alexey Dosovitskiy}, \bibinfo{person}{Lucas Beyer}, \bibinfo{person}{Alexander Kolesnikov}, \bibinfo{person}{Dirk Weissenborn}, \bibinfo{person}{Xiaohua Zhai}, \bibinfo{person}{Thomas Unterthiner}, \bibinfo{person}{Mostafa Dehghani}, \bibinfo{person}{Matthias Minderer}, \bibinfo{person}{Georg Heigold}, \bibinfo{person}{Sylvain Gelly}, {et~al\mbox{.}}} \bibinfo{year}{2020}\natexlab{}.
\newblock \showarticletitle{An image is worth 16x16 words: Transformers for image recognition at scale}.
\newblock \bibinfo{journal}{\emph{arXiv preprint arXiv:2010.11929}} (\bibinfo{year}{2020}).
\newblock


\bibitem[Fallah et~al\mbox{.}(2020)]%
        {fallah2020personalized}
\bibfield{author}{\bibinfo{person}{Alireza Fallah}, \bibinfo{person}{Aryan Mokhtari}, {and} \bibinfo{person}{Asuman Ozdaglar}.} \bibinfo{year}{2020}\natexlab{}.
\newblock \showarticletitle{Personalized federated learning with theoretical guarantees: A model-agnostic meta-learning approach}.
\newblock \bibinfo{journal}{\emph{Advances in Neural Information Processing Systems}}  \bibinfo{volume}{33} (\bibinfo{year}{2020}), \bibinfo{pages}{3557--3568}.
\newblock


\bibitem[Fang et~al\mbox{.}(2019)]%
        {fang2019gstnet}
\bibfield{author}{\bibinfo{person}{Shen Fang}, \bibinfo{person}{Qi Zhang}, \bibinfo{person}{Gaofeng Meng}, \bibinfo{person}{Shiming Xiang}, {and} \bibinfo{person}{Chunhong Pan}.} \bibinfo{year}{2019}\natexlab{}.
\newblock \showarticletitle{GSTNet: Global spatial-temporal network for traffic flow prediction.}. In \bibinfo{booktitle}{\emph{IJCAI}}. \bibinfo{pages}{2286--2293}.
\newblock


\bibitem[Guo et~al\mbox{.}(2023)]%
        {guo2023pfedprompt}
\bibfield{author}{\bibinfo{person}{Tao Guo}, \bibinfo{person}{Song Guo}, {and} \bibinfo{person}{Junxiao Wang}.} \bibinfo{year}{2023}\natexlab{}.
\newblock \showarticletitle{pFedPrompt: Learning Personalized Prompt for Vision-Language Models in Federated Learning}. In \bibinfo{booktitle}{\emph{Proceedings of the ACM Web Conference}}. \bibinfo{pages}{1364--1374}.
\newblock


\bibitem[He et~al\mbox{.}(2016)]%
        {he2016deep}
\bibfield{author}{\bibinfo{person}{Kaiming He}, \bibinfo{person}{Xiangyu Zhang}, \bibinfo{person}{Shaoqing Ren}, {and} \bibinfo{person}{Jian Sun}.} \bibinfo{year}{2016}\natexlab{}.
\newblock \showarticletitle{Deep residual learning for image recognition}. In \bibinfo{booktitle}{\emph{Proceedings of the IEEE Conference on Computer Vision and Pattern Recognition}}. \bibinfo{pages}{770--778}.
\newblock


\bibitem[Hu et~al\mbox{.}(2023)]%
        {hu2023vl}
\bibfield{author}{\bibinfo{person}{Zi-Yuan Hu}, \bibinfo{person}{Yanyang Li}, \bibinfo{person}{Michael~R Lyu}, {and} \bibinfo{person}{Liwei Wang}.} \bibinfo{year}{2023}\natexlab{}.
\newblock \showarticletitle{Vl-pet: Vision-and-language parameter-efficient tuning via granularity control}. In \bibinfo{booktitle}{\emph{Proceedings of the IEEE/CVF International Conference on Computer Vision}}. \bibinfo{pages}{3010--3020}.
\newblock


\bibitem[Jia et~al\mbox{.}(2022)]%
        {jia2022visual}
\bibfield{author}{\bibinfo{person}{Menglin Jia}, \bibinfo{person}{Luming Tang}, \bibinfo{person}{Bor-Chun Chen}, \bibinfo{person}{Claire Cardie}, \bibinfo{person}{Serge Belongie}, \bibinfo{person}{Bharath Hariharan}, {and} \bibinfo{person}{Ser-Nam Lim}.} \bibinfo{year}{2022}\natexlab{}.
\newblock \showarticletitle{Visual prompt tuning}. In \bibinfo{booktitle}{\emph{European Conference on Computer Vision}}. Springer, \bibinfo{pages}{709--727}.
\newblock


\bibitem[Kang et~al\mbox{.}(2023a)]%
        {kang2023grounding}
\bibfield{author}{\bibinfo{person}{Yan Kang}, \bibinfo{person}{Tao Fan}, \bibinfo{person}{Hanlin Gu}, \bibinfo{person}{Lixin Fan}, {and} \bibinfo{person}{Qiang Yang}.} \bibinfo{year}{2023}\natexlab{a}.
\newblock \showarticletitle{Grounding foundation models through federated transfer learning: A general framework}.
\newblock \bibinfo{journal}{\emph{arXiv preprint arXiv:2311.17431}} (\bibinfo{year}{2023}).
\newblock


\bibitem[Kang et~al\mbox{.}(2023b)]%
        {kang2023optimizing}
\bibfield{author}{\bibinfo{person}{Yan Kang}, \bibinfo{person}{Hanlin Gu}, \bibinfo{person}{Xingxing Tang}, \bibinfo{person}{Yuanqin He}, \bibinfo{person}{Yuzhu Zhang}, \bibinfo{person}{Jinnan He}, \bibinfo{person}{Yuxing Han}, \bibinfo{person}{Lixin Fan}, \bibinfo{person}{Kai Chen}, {and} \bibinfo{person}{Qiang Yang}.} \bibinfo{year}{2023}\natexlab{b}.
\newblock \showarticletitle{Optimizing privacy, utility and efficiency in constrained multi-objective federated learning}.
\newblock \bibinfo{journal}{\emph{arXiv preprint arXiv:2305.00312}} (\bibinfo{year}{2023}).
\newblock


\bibitem[Khan et~al\mbox{.}(2023)]%
        {khan2023introducing}
\bibfield{author}{\bibinfo{person}{Muhammad Gul Zain~Ali Khan}, \bibinfo{person}{Muhammad~Ferjad Naeem}, \bibinfo{person}{Luc Van~Gool}, \bibinfo{person}{Didier Stricker}, \bibinfo{person}{Federico Tombari}, {and} \bibinfo{person}{Muhammad~Zeshan Afzal}.} \bibinfo{year}{2023}\natexlab{}.
\newblock \showarticletitle{Introducing language guidance in prompt-based continual learning}. In \bibinfo{booktitle}{\emph{Proceedings of the IEEE/CVF International Conference on Computer Vision}}. \bibinfo{pages}{11463--11473}.
\newblock


\bibitem[Kirkpatrick et~al\mbox{.}(2017)]%
        {kirkpatrick2017overcoming}
\bibfield{author}{\bibinfo{person}{James Kirkpatrick}, \bibinfo{person}{Razvan Pascanu}, \bibinfo{person}{Neil Rabinowitz}, \bibinfo{person}{Joel Veness}, \bibinfo{person}{Guillaume Desjardins}, \bibinfo{person}{Andrei~A Rusu}, \bibinfo{person}{Kieran Milan}, \bibinfo{person}{John Quan}, \bibinfo{person}{Tiago Ramalho}, \bibinfo{person}{Agnieszka Grabska-Barwinska}, {et~al\mbox{.}}} \bibinfo{year}{2017}\natexlab{}.
\newblock \showarticletitle{Overcoming catastrophic forgetting in neural networks}.
\newblock \bibinfo{journal}{\emph{Proceedings of the National Academy of Sciences}} \bibinfo{volume}{114}, \bibinfo{number}{13} (\bibinfo{year}{2017}), \bibinfo{pages}{3521--3526}.
\newblock


\bibitem[Krizhevsky et~al\mbox{.}(2009)]%
        {krizhevsky2009learning}
\bibfield{author}{\bibinfo{person}{Alex Krizhevsky}, \bibinfo{person}{Geoffrey Hinton}, {et~al\mbox{.}}} \bibinfo{year}{2009}\natexlab{}.
\newblock \showarticletitle{Learning multiple layers of features from tiny images}.
\newblock  (\bibinfo{year}{2009}).
\newblock


\bibitem[Lester et~al\mbox{.}(2021)]%
        {lester2021power}
\bibfield{author}{\bibinfo{person}{Brian Lester}, \bibinfo{person}{Rami Al-Rfou}, {and} \bibinfo{person}{Noah Constant}.} \bibinfo{year}{2021}\natexlab{}.
\newblock \showarticletitle{The Power of Scale for Parameter-Efficient Prompt Tuning}. In \bibinfo{booktitle}{\emph{Proceedings of the 2021 Conference on Empirical Methods in Natural Language Processing}}. \bibinfo{pages}{3045--3059}.
\newblock


\bibitem[Li et~al\mbox{.}(2023b)]%
        {li2023cl}
\bibfield{author}{\bibinfo{person}{Miaomiao Li}, \bibinfo{person}{Jiaqi Zhu}, \bibinfo{person}{Xin Yang}, \bibinfo{person}{Yi Yang}, \bibinfo{person}{Qiang Gao}, {and} \bibinfo{person}{Hongan Wang}.} \bibinfo{year}{2023}\natexlab{b}.
\newblock \showarticletitle{CL-WSTC: Continual Learning for Weakly Supervised Text Classification on the Internet}. In \bibinfo{booktitle}{\emph{Proceedings of the ACM Web Conference}}. \bibinfo{pages}{1489--1499}.
\newblock


\bibitem[Li et~al\mbox{.}(2021)]%
        {li2021ditto}
\bibfield{author}{\bibinfo{person}{Tian Li}, \bibinfo{person}{Shengyuan Hu}, \bibinfo{person}{Ahmad Beirami}, {and} \bibinfo{person}{Virginia Smith}.} \bibinfo{year}{2021}\natexlab{}.
\newblock \showarticletitle{Ditto: Fair and robust federated learning through personalization}. In \bibinfo{booktitle}{\emph{International Conference on Machine Learning}}. PMLR, \bibinfo{pages}{6357--6368}.
\newblock


\bibitem[Li et~al\mbox{.}(2020)]%
        {li2020federated}
\bibfield{author}{\bibinfo{person}{Tian Li}, \bibinfo{person}{Anit~Kumar Sahu}, \bibinfo{person}{Manzil Zaheer}, \bibinfo{person}{Maziar Sanjabi}, \bibinfo{person}{Ameet Talwalkar}, {and} \bibinfo{person}{Virginia Smith}.} \bibinfo{year}{2020}\natexlab{}.
\newblock \showarticletitle{Federated optimization in heterogeneous networks}.
\newblock \bibinfo{journal}{\emph{Proceedings of Machine Learning and Systems}}  \bibinfo{volume}{2} (\bibinfo{year}{2020}), \bibinfo{pages}{429--450}.
\newblock


\bibitem[Li et~al\mbox{.}(2024)]%
        {li2024towards}
\bibfield{author}{\bibinfo{person}{Yichen Li}, \bibinfo{person}{Qunwei Li}, \bibinfo{person}{Haozhao Wang}, \bibinfo{person}{Ruixuan Li}, \bibinfo{person}{Wenliang Zhong}, {and} \bibinfo{person}{Guannan Zhang}.} \bibinfo{year}{2024}\natexlab{}.
\newblock \showarticletitle{Towards Efficient Replay in Federated Incremental Learning}.
\newblock \bibinfo{journal}{\emph{arXiv preprint arXiv:2403.05890}} (\bibinfo{year}{2024}).
\newblock


\bibitem[Li et~al\mbox{.}(2023a)]%
        {li2023learning}
\bibfield{author}{\bibinfo{person}{Yujie Li}, \bibinfo{person}{Xin Yang}, \bibinfo{person}{Hao Wang}, \bibinfo{person}{Xiangkun Wang}, {and} \bibinfo{person}{Tianrui Li}.} \bibinfo{year}{2023}\natexlab{a}.
\newblock \showarticletitle{Learning to Prompt Knowledge Transfer for Open-World Continual Learning}.
\newblock \bibinfo{journal}{\emph{arXiv preprint arXiv:2312.14990}} (\bibinfo{year}{2023}).
\newblock


\bibitem[Liu et~al\mbox{.}(2020)]%
        {liu2020client}
\bibfield{author}{\bibinfo{person}{Lumin Liu}, \bibinfo{person}{Jun Zhang}, \bibinfo{person}{SH Song}, {and} \bibinfo{person}{Khaled~B Letaief}.} \bibinfo{year}{2020}\natexlab{}.
\newblock \showarticletitle{Client-edge-cloud hierarchical federated learning}. In \bibinfo{booktitle}{\emph{ICC 2020-2020 IEEE International Conference on Communications}}. IEEE, \bibinfo{pages}{1--6}.
\newblock


\bibitem[Liu et~al\mbox{.}(2024)]%
        {liu2024vfl}
\bibfield{author}{\bibinfo{person}{Yang Liu}, \bibinfo{person}{Yan Kang}, \bibinfo{person}{Tianyuan Zou}, \bibinfo{person}{Yanhong Pu}, \bibinfo{person}{Yuanqin He}, \bibinfo{person}{Xiaozhou Ye}, \bibinfo{person}{Ye Ouyang}, \bibinfo{person}{Ya-Qin Zhang}, {and} \bibinfo{person}{Qiang Yang}.} \bibinfo{year}{2024}\natexlab{}.
\newblock \showarticletitle{Vertical Federated Learning: Concepts, Advances, and Challenges}.
\newblock \bibinfo{journal}{\emph{IEEE Transactions on Knowledge and Data Engineering}} (\bibinfo{year}{2024}).
\newblock
\urldef\tempurl%
\url{https://doi.org/10.1109/TKDE.2024.3352628}
\showDOI{\tempurl}


\bibitem[Ma et~al\mbox{.}(2022)]%
        {ma2022layer}
\bibfield{author}{\bibinfo{person}{Xiaosong Ma}, \bibinfo{person}{Jie Zhang}, \bibinfo{person}{Song Guo}, {and} \bibinfo{person}{Wenchao Xu}.} \bibinfo{year}{2022}\natexlab{}.
\newblock \showarticletitle{Layer-wised model aggregation for personalized federated learning}. In \bibinfo{booktitle}{\emph{Proceedings of the IEEE/CVF Conference on Computer Vision and Pattern Recognition}}. \bibinfo{pages}{10092--10101}.
\newblock


\bibitem[Mai et~al\mbox{.}(2022)]%
        {mai2022online}
\bibfield{author}{\bibinfo{person}{Zheda Mai}, \bibinfo{person}{Ruiwen Li}, \bibinfo{person}{Jihwan Jeong}, \bibinfo{person}{David Quispe}, \bibinfo{person}{Hyunwoo Kim}, {and} \bibinfo{person}{Scott Sanner}.} \bibinfo{year}{2022}\natexlab{}.
\newblock \showarticletitle{Online continual learning in image classification: An empirical survey}.
\newblock \bibinfo{journal}{\emph{Neurocomputing}}  \bibinfo{volume}{469} (\bibinfo{year}{2022}), \bibinfo{pages}{28--51}.
\newblock


\bibitem[Masana et~al\mbox{.}(2022)]%
        {masana2022class}
\bibfield{author}{\bibinfo{person}{Marc Masana}, \bibinfo{person}{Xialei Liu}, \bibinfo{person}{Bart{\l}omiej Twardowski}, \bibinfo{person}{Mikel Menta}, \bibinfo{person}{Andrew~D Bagdanov}, {and} \bibinfo{person}{Joost Van De~Weijer}.} \bibinfo{year}{2022}\natexlab{}.
\newblock \showarticletitle{Class-incremental learning: survey and performance evaluation on image classification}.
\newblock \bibinfo{journal}{\emph{IEEE Transactions on Pattern Analysis and Machine Intelligence}} \bibinfo{volume}{45}, \bibinfo{number}{5} (\bibinfo{year}{2022}), \bibinfo{pages}{5513--5533}.
\newblock


\bibitem[McMahan et~al\mbox{.}(2017)]%
        {mcmahan2017communication}
\bibfield{author}{\bibinfo{person}{Brendan McMahan}, \bibinfo{person}{Eider Moore}, \bibinfo{person}{Daniel Ramage}, \bibinfo{person}{Seth Hampson}, {and} \bibinfo{person}{Blaise~Aguera y Arcas}.} \bibinfo{year}{2017}\natexlab{}.
\newblock \showarticletitle{Communication-efficient learning of deep networks from decentralized data}. In \bibinfo{booktitle}{\emph{Artificial Intelligence and Statistics}}. PMLR, \bibinfo{pages}{1273--1282}.
\newblock


\bibitem[Pan et~al\mbox{.}(2021)]%
        {pan2021privacy}
\bibfield{author}{\bibinfo{person}{Zijie Pan}, \bibinfo{person}{Li Hu}, \bibinfo{person}{Weixuan Tang}, \bibinfo{person}{Jin Li}, \bibinfo{person}{Yi He}, {and} \bibinfo{person}{Zheli Liu}.} \bibinfo{year}{2021}\natexlab{}.
\newblock \showarticletitle{Privacy-preserving multi-granular federated neural architecture search a general framework}.
\newblock \bibinfo{journal}{\emph{IEEE Transactions on Knowledge and Data Engineering}} (\bibinfo{year}{2021}).
\newblock


\bibitem[Pillutla et~al\mbox{.}(2022)]%
        {pillutla2022federated}
\bibfield{author}{\bibinfo{person}{Krishna Pillutla}, \bibinfo{person}{Kshitiz Malik}, \bibinfo{person}{Abdel-Rahman Mohamed}, \bibinfo{person}{Mike Rabbat}, \bibinfo{person}{Maziar Sanjabi}, {and} \bibinfo{person}{Lin Xiao}.} \bibinfo{year}{2022}\natexlab{}.
\newblock \showarticletitle{Federated learning with partial model personalization}. In \bibinfo{booktitle}{\emph{International Conference on Machine Learning}}. PMLR, \bibinfo{pages}{17716--17758}.
\newblock


\bibitem[Smith et~al\mbox{.}(2023)]%
        {smith2023coda}
\bibfield{author}{\bibinfo{person}{James~Seale Smith}, \bibinfo{person}{Leonid Karlinsky}, \bibinfo{person}{Vyshnavi Gutta}, \bibinfo{person}{Paola Cascante-Bonilla}, \bibinfo{person}{Donghyun Kim}, \bibinfo{person}{Assaf Arbelle}, \bibinfo{person}{Rameswar Panda}, \bibinfo{person}{Rogerio Feris}, {and} \bibinfo{person}{Zsolt Kira}.} \bibinfo{year}{2023}\natexlab{}.
\newblock \showarticletitle{CODA-Prompt: COntinual Decomposed Attention-based Prompting for Rehearsal-Free Continual Learning}. In \bibinfo{booktitle}{\emph{Proceedings of the IEEE/CVF Conference on Computer Vision and Pattern Recognition}}. \bibinfo{pages}{11909--11919}.
\newblock


\bibitem[Sun et~al\mbox{.}(2021)]%
        {sun2021partialfed}
\bibfield{author}{\bibinfo{person}{Benyuan Sun}, \bibinfo{person}{Hongxing Huo}, \bibinfo{person}{Yi Yang}, {and} \bibinfo{person}{Bo Bai}.} \bibinfo{year}{2021}\natexlab{}.
\newblock \showarticletitle{Partialfed: Cross-domain personalized federated learning via partial initialization}.
\newblock \bibinfo{journal}{\emph{Advances in Neural Information Processing Systems}}  \bibinfo{volume}{34} (\bibinfo{year}{2021}), \bibinfo{pages}{23309--23320}.
\newblock


\bibitem[T~Dinh et~al\mbox{.}(2020)]%
        {t2020personalized}
\bibfield{author}{\bibinfo{person}{Canh T~Dinh}, \bibinfo{person}{Nguyen Tran}, {and} \bibinfo{person}{Josh Nguyen}.} \bibinfo{year}{2020}\natexlab{}.
\newblock \showarticletitle{Personalized federated learning with Moreau envelopes}.
\newblock \bibinfo{journal}{\emph{Advances in Neural Information Processing Systems}}  \bibinfo{volume}{33} (\bibinfo{year}{2020}), \bibinfo{pages}{21394--21405}.
\newblock


\bibitem[Tan et~al\mbox{.}(2022)]%
        {tan2022towards}
\bibfield{author}{\bibinfo{person}{Alysa~Ziying Tan}, \bibinfo{person}{Han Yu}, \bibinfo{person}{Lizhen Cui}, {and} \bibinfo{person}{Qiang Yang}.} \bibinfo{year}{2022}\natexlab{}.
\newblock \showarticletitle{Towards personalized federated learning}.
\newblock \bibinfo{journal}{\emph{IEEE Transactions on Neural Networks and Learning Systems}} (\bibinfo{year}{2022}).
\newblock


\bibitem[Vahidian et~al\mbox{.}(2023)]%
        {vahidian2023efficient}
\bibfield{author}{\bibinfo{person}{Saeed Vahidian}, \bibinfo{person}{Mahdi Morafah}, \bibinfo{person}{Weijia Wang}, \bibinfo{person}{Vyacheslav Kungurtsev}, \bibinfo{person}{Chen Chen}, \bibinfo{person}{Mubarak Shah}, {and} \bibinfo{person}{Bill Lin}.} \bibinfo{year}{2023}\natexlab{}.
\newblock \showarticletitle{Efficient distribution similarity identification in clustered federated learning via principal angles between client data subspaces}. In \bibinfo{booktitle}{\emph{Proceedings of the AAAI Conference on Artificial Intelligence}}, Vol.~\bibinfo{volume}{37}. \bibinfo{pages}{10043--10052}.
\newblock


\bibitem[Vaswani et~al\mbox{.}(2017)]%
        {vaswani2017attention}
\bibfield{author}{\bibinfo{person}{Ashish Vaswani}, \bibinfo{person}{Noam Shazeer}, \bibinfo{person}{Niki Parmar}, \bibinfo{person}{Jakob Uszkoreit}, \bibinfo{person}{Llion Jones}, \bibinfo{person}{Aidan~N Gomez}, \bibinfo{person}{{\L}ukasz Kaiser}, {and} \bibinfo{person}{Illia Polosukhin}.} \bibinfo{year}{2017}\natexlab{}.
\newblock \showarticletitle{Attention is all you need}.
\newblock \bibinfo{journal}{\emph{Advances in Neural Information Processing Systems}}  \bibinfo{volume}{30} (\bibinfo{year}{2017}).
\newblock


\bibitem[Wang et~al\mbox{.}(2022a)]%
        {wang2022dualprompt}
\bibfield{author}{\bibinfo{person}{Zifeng Wang}, \bibinfo{person}{Zizhao Zhang}, \bibinfo{person}{Sayna Ebrahimi}, \bibinfo{person}{Ruoxi Sun}, \bibinfo{person}{Han Zhang}, \bibinfo{person}{Chen-Yu Lee}, \bibinfo{person}{Xiaoqi Ren}, \bibinfo{person}{Guolong Su}, \bibinfo{person}{Vincent Perot}, \bibinfo{person}{Jennifer Dy}, {et~al\mbox{.}}} \bibinfo{year}{2022}\natexlab{a}.
\newblock \showarticletitle{Dualprompt: Complementary prompting for rehearsal-free continual learning}. In \bibinfo{booktitle}{\emph{European Conference on Computer Vision}}. Springer, \bibinfo{pages}{631--648}.
\newblock


\bibitem[Wang et~al\mbox{.}(2022b)]%
        {wang2022learning}
\bibfield{author}{\bibinfo{person}{Zifeng Wang}, \bibinfo{person}{Zizhao Zhang}, \bibinfo{person}{Chen-Yu Lee}, \bibinfo{person}{Han Zhang}, \bibinfo{person}{Ruoxi Sun}, \bibinfo{person}{Xiaoqi Ren}, \bibinfo{person}{Guolong Su}, \bibinfo{person}{Vincent Perot}, \bibinfo{person}{Jennifer Dy}, {and} \bibinfo{person}{Tomas Pfister}.} \bibinfo{year}{2022}\natexlab{b}.
\newblock \showarticletitle{Learning to prompt for continual learning}. In \bibinfo{booktitle}{\emph{Proceedings of the IEEE/CVF Conference on Computer Vision and Pattern Recognition}}. \bibinfo{pages}{139--149}.
\newblock


\bibitem[Xiao et~al\mbox{.}(2018)]%
        {xiao2018group}
\bibfield{author}{\bibinfo{person}{Hao Xiao}, \bibinfo{person}{Weiyao Lin}, \bibinfo{person}{Bin Sheng}, \bibinfo{person}{Ke Lu}, \bibinfo{person}{Junchi Yan}, \bibinfo{person}{Jingdong Wang}, \bibinfo{person}{Errui Ding}, \bibinfo{person}{Yihao Zhang}, {and} \bibinfo{person}{Hongkai Xiong}.} \bibinfo{year}{2018}\natexlab{}.
\newblock \showarticletitle{Group re-identification: Leveraging and integrating multi-grain information}. In \bibinfo{booktitle}{\emph{Proceedings of the 26th ACM International Conference on Multimedia}}. \bibinfo{pages}{192--200}.
\newblock


\bibitem[Yang et~al\mbox{.}(2023)]%
        {yang2023efficient}
\bibfield{author}{\bibinfo{person}{Fu-En Yang}, \bibinfo{person}{Chien-Yi Wang}, {and} \bibinfo{person}{Yu-Chiang~Frank Wang}.} \bibinfo{year}{2023}\natexlab{}.
\newblock \showarticletitle{Efficient model personalization in federated learning via client-specific prompt generation}. In \bibinfo{booktitle}{\emph{Proceedings of the IEEE/CVF International Conference on Computer Vision}}. \bibinfo{pages}{19159--19168}.
\newblock


\bibitem[Yang et~al\mbox{.}(2020)]%
        {yang2020fedsteg}
\bibfield{author}{\bibinfo{person}{Hongwei Yang}, \bibinfo{person}{Hui He}, \bibinfo{person}{Weizhe Zhang}, {and} \bibinfo{person}{Xiaochun Cao}.} \bibinfo{year}{2020}\natexlab{}.
\newblock \showarticletitle{FedSteg: A federated transfer learning framework for secure image steganalysis}.
\newblock \bibinfo{journal}{\emph{IEEE Transactions on Network Science and Engineering}} \bibinfo{volume}{8}, \bibinfo{number}{2} (\bibinfo{year}{2020}), \bibinfo{pages}{1084--1094}.
\newblock


\bibitem[Yang et~al\mbox{.}(2022a)]%
        {yang2022temporal}
\bibfield{author}{\bibinfo{person}{Xin Yang}, \bibinfo{person}{Yujie Li}, \bibinfo{person}{Qiuke Li}, \bibinfo{person}{Dun Liu}, {and} \bibinfo{person}{Tianrui Li}.} \bibinfo{year}{2022}\natexlab{a}.
\newblock \showarticletitle{Temporal-spatial three-way granular computing for dynamic text sentiment classification}.
\newblock \bibinfo{journal}{\emph{Information Sciences}}  \bibinfo{volume}{596} (\bibinfo{year}{2022}), \bibinfo{pages}{551--566}.
\newblock


\bibitem[Yang et~al\mbox{.}(2022b)]%
        {yang2022three}
\bibfield{author}{\bibinfo{person}{Xin Yang}, \bibinfo{person}{Yujie Li}, \bibinfo{person}{Dan Meng}, \bibinfo{person}{Yuxuan Yang}, \bibinfo{person}{Dun Liu}, {and} \bibinfo{person}{Tianrui Li}.} \bibinfo{year}{2022}\natexlab{b}.
\newblock \showarticletitle{Three-way multi-granularity learning towards open topic classification}.
\newblock \bibinfo{journal}{\emph{Information Sciences}}  \bibinfo{volume}{585} (\bibinfo{year}{2022}), \bibinfo{pages}{41--57}.
\newblock


\bibitem[Yang et~al\mbox{.}(2024)]%
        {yang2023federated}
\bibfield{author}{\bibinfo{person}{Xin Yang}, \bibinfo{person}{Hao Yu}, \bibinfo{person}{Xin Gao}, \bibinfo{person}{Hao Wang}, \bibinfo{person}{Junbo Zhang}, {and} \bibinfo{person}{Tianrui Li}.} \bibinfo{year}{2024}\natexlab{}.
\newblock \showarticletitle{Federated Continual Learning via Knowledge Fusion: A Survey}.
\newblock \bibinfo{journal}{\emph{IEEE Transactions on Knowledge and Data Engineering}} (\bibinfo{year}{2024}).
\newblock


\bibitem[Zhang et~al\mbox{.}(2023)]%
        {zhang2023fedcp}
\bibfield{author}{\bibinfo{person}{Jianqing Zhang}, \bibinfo{person}{Yang Hua}, \bibinfo{person}{Hao Wang}, \bibinfo{person}{Tao Song}, \bibinfo{person}{Zhengui Xue}, \bibinfo{person}{Ruhui Ma}, {and} \bibinfo{person}{Haibing Guan}.} \bibinfo{year}{2023}\natexlab{}.
\newblock \showarticletitle{Fedcp: Separating feature information for personalized federated learning via conditional policy}. In \bibinfo{booktitle}{\emph{Proceedings of the 29th ACM SIGKDD Conference on Knowledge Discovery and Data Mining}}. \bibinfo{pages}{3249--3261}.
\newblock


\bibitem[Zhou et~al\mbox{.}(2022)]%
        {zhou2022learning}
\bibfield{author}{\bibinfo{person}{Kaiyang Zhou}, \bibinfo{person}{Jingkang Yang}, \bibinfo{person}{Chen~Change Loy}, {and} \bibinfo{person}{Ziwei Liu}.} \bibinfo{year}{2022}\natexlab{}.
\newblock \showarticletitle{Learning to prompt for vision-language models}.
\newblock \bibinfo{journal}{\emph{International Journal of Computer Vision}} \bibinfo{volume}{130}, \bibinfo{number}{9} (\bibinfo{year}{2022}), \bibinfo{pages}{2337--2348}.
\newblock


\bibitem[Zhou et~al\mbox{.}(2020)]%
        {zhou2020foresee}
\bibfield{author}{\bibinfo{person}{Zhengyang Zhou}, \bibinfo{person}{Yang Wang}, \bibinfo{person}{Xike Xie}, \bibinfo{person}{Lianliang Chen}, {and} \bibinfo{person}{Chaochao Zhu}.} \bibinfo{year}{2020}\natexlab{}.
\newblock \showarticletitle{Foresee urban sparse traffic accidents: A spatiotemporal multi-granularity perspective}.
\newblock \bibinfo{journal}{\emph{IEEE Transactions on Knowledge and Data Engineering}} \bibinfo{volume}{34}, \bibinfo{number}{8} (\bibinfo{year}{2020}), \bibinfo{pages}{3786--3799}.
\newblock


\end{thebibliography}
